%% file: main.tex
\documentclass[sigconf]{acmart}
\usepackage{url}  
\usepackage{graphicx}
\usepackage{color}
\usepackage{amsmath}
\usepackage{natbib}
\usepackage{algorithm}
\usepackage{algorithmic}
\usepackage{calc}
\usepackage{siunitx}
\usepackage{bm}

\newtheorem{definition}{Definition}

\DeclareMathOperator{\E}{\mathbb{E}}

\newcommand{\method}{HeroCon}

\ExplSyntaxOn
\newcommand\latinabbrev[1]{
  \peek_meaning:NTF . {
    #1\@}%
  { \peek_catcode:NTF a {
      #1.\@ }%
    {#1.\@}}}
\ExplSyntaxOff

\def\eg {\latinabbrev{e.g}}

\def\ie{\latinabbrev{i.e}}

\ccsdesc[500]{Computing methodologies~Supervised learning}


\copyrightyear{2022}
\acmYear{2022}
\setcopyright{acmcopyright}
\acmConference[KDD '22] {Proceedings of the 28th ACM SIGKDD Conference on Knowledge Discovery and Data Mining}{August 14--18, 2022}{Washington, DC, USA.}
\acmBooktitle{Proceedings of the 28th ACM SIGKDD Conference on Knowledge Discovery and Data Mining (KDD '22), August 14--18, 2022, Washington, DC, USA}
\acmPrice{15.00}
\acmISBN{978-1-4503-9385-0/22/08}
\acmDOI{10.1145/3534678.3539311}

\begin{document}

\title{Contrastive Learning with Complex Heterogeneity}

\author{Lecheng Zheng}
\affiliation{%
  \institution{University of Illinois at Urbana-Champaign}
  \country{Illinois, USA}}
\email{lecheng4@illinois.edu}

\author{Jinjun Xiong}
\affiliation{%
  \institution{University at Buffalo}
  \country{New York, USA}}
\email{jinjun@buffalo.edu}

\author{Yada Zhu}
\affiliation{%
  \institution{MIT-IBM Watson AI Lab, IBM Research}
  \country{New York, USA}}
\email{yzhu@us.ibm.com}

\author{Jingrui He}
\affiliation{%
  \institution{University of Illinois at Urbana-Champaign}
  \country{Illinois, USA}}
\email{jingrui@illinois.edu}

\renewcommand{\shortauthors}{Lecheng Zheng et al.}

\input{abstract.tex}
\date{}
\maketitle
\input{Introduction}

\input{Related_Work}

\input{Algorithm}

\input{Experiment}

\input{Conclusion}
\section*{Acknowledgment}
This work is supported by National Science Foundation under Award No. IIS-1947203, IIS-2117902, IIS-2137468, the C3.ai Digital Transformation Institute, MIT-IBM Watson AI Lab, and IBM-ILLINOIS Center for Cognitive Computing Systems Research (C3SR) -- a research collaboration as part of the IBM AI Horizons Network. The views and conclusions are those of the authors and should not be interpreted as representing the official policies of the funding agencies or the government.

\bibliographystyle{ACM-Reference-Format}
{\small
\bibliography{reference}
}
\newpage
\appendix
\balance

\input{proof}

\end{document}

%% file: abstract.tex
\begin{abstract}
    With the advent of big data across multiple high-impact applications, we are often facing the challenge of complex heterogeneity. The newly collected data usually consist of multiple modalities and are characterized with multiple labels, thus exhibiting the co-existence of multiple types of heterogeneity. Although state-of-the-art techniques are good at modeling the complex heterogeneity with sufficient label information, such label information can be quite expensive to obtain in real applications. Recently, researchers pay great attention to contrastive learning due to its prominent performance by utilizing rich unlabeled data. However, existing work on contrastive learning is not able to address the problem of false negative pairs, i.e., some `negative' pairs may have similar representations if they have the same label.
    To overcome the issues, in this paper, we propose a unified heterogeneous learning framework, which combines both the weighted unsupervised contrastive loss and the weighted supervised contrastive loss to model multiple types of heterogeneity. We first provide a theoretical analysis showing that the vanilla contrastive learning loss easily leads to the sub-optimal solution in the presence of false negative pairs, whereas the proposed weighted loss could automatically adjust the weight based on the similarity of the learned representations to mitigate this issue. 
    Experimental results on real-world data sets demonstrate the effectiveness and the efficiency of the proposed framework modeling multiple types of heterogeneity.
\end{abstract}

%% file: Introduction.tex
\section{Introduction}
Recent years have witnessed the surge of big data. According to a report published in Forbes\footnote{\url{https://www.forbes.com/sites/gilpress/2020/01/06/6-predictions-about-data-in-2020-and-the-coming-decade/?sh=3214c68f4fc3}}, the amount of newly created data in the past two years had increased by more than two trillion gigabytes. One major characteristic of big data is variety or heterogeneity. Furthermore, many high-impact applications exhibit complex heterogeneity or the co-existence of multiple types of data heterogeneity. For example, in social media, a post may consist of both image data and text data, i.e., view heterogeneity, and it can be assigned multiple tags based on the content, i.e., label heterogeneity; in the financial domain, the stock related data may be collected from multiple sources (\eg, financial reports, weather, and news)\cite{zhou2020domain}, and the corresponding labels may not only include the stock price but also the price trend or volatility.
To model such complex heterogeneity, heterogeneous learning has been studied for decades. Initial efforts focused on shallow machine learning algorithms modeling single heterogeneity (\eg,~\cite{zhang2010multi, nigam2000analyzing, DBLP:conf/icml/ZhouB07, DBLP:conf/nips/ZhouCY11, DBLP:conf/icml/KimX10}), or dual heterogeneity (e.g.,~\cite{he2011graphbased, DBLP:conf/iccv/HongMPT13,DBLP:conf/aaai/LuoTXLX13}). More recently, many researchers started exploring deep neural network based algorithms~\cite{DBLP:conf/cvpr/LuKZCJF17, DBLP:conf/cvpr/MisraSGH16, DBLP:journals/corr/MaoXYWY14, DBLP:conf/sdm/ZhengCH19, zheng2021deep}, which achieved state-of-the-art performance in many scenarios.
However, most (if not all) of these algorithms rely on large amount of label information to build accurate models, which can be expensive and time-consuming to obtain in real applications. In other words, if applied to a data set consisting of large amount of unlabeled data and only a small percentage of labeled data, these algorithms may only lead to sub-optimal performance.

\begin{figure}
\begin{center}
\includegraphics[width=0.70\linewidth]{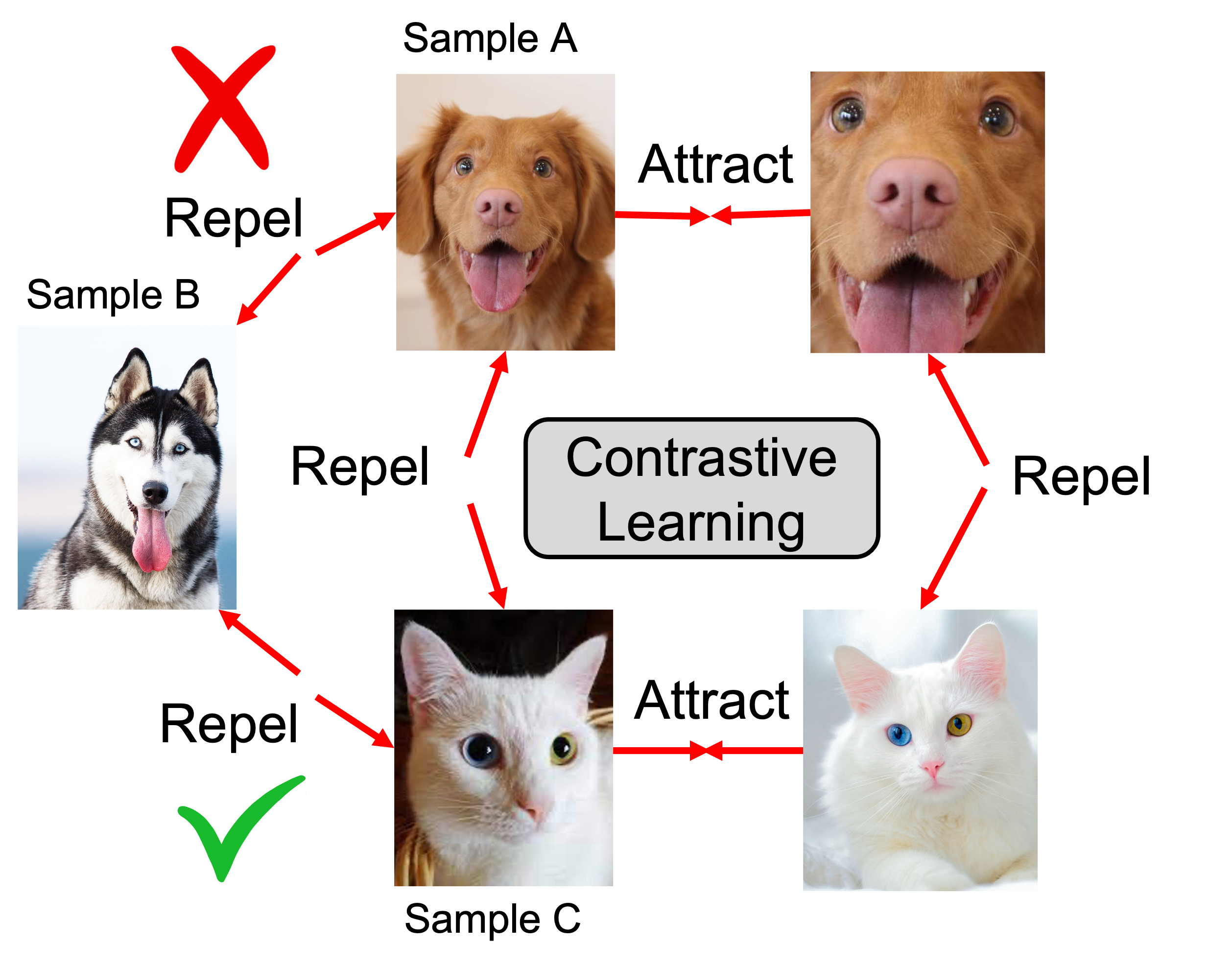}
\end{center}
\vspace{-0.4cm}
\caption{Issue of the vanilla contrastive learning}
\label{fig_motivation}
\vspace{-0.5cm}
\end{figure}

In recent years, researchers pay great attention to contrastive learning due to its prominent performance leveraging the rich unlabeled data to improve the performance~\cite{oord2018representation, song2020multi, chuang2020debiased, khosla2020supervised, tian2019contrastive, chen2020simple}. However, it fails to address the challenging scenario where some `negative' pairs may have similar or even identical representations. For example, given a binary classification data set in Figure~\ref{fig_motivation}, vanilla contrastive learning framework aims to learn the hidden representation by contrasting the representation of one sample (\eg, \textit{Sample A}) with the the representation of another sample (\eg, \textit{Sample B}). However, if \textit{Sample B} has the same label as \textit{Sample A}, then both samples tend to have similar hidden representations. In this case, vanilla contrastive learning may lead to a sub-optimal solution because contrastive learning loss pushes the hidden representation of \textit{Sample A} away from that of \textit{Sample B}.
The current existing work~\cite{chuang2020debiased} imposes the identical weights on the negative pairs to tighten the lower bound of the mutual information but it still fails to alleviate the aforementioned negative impact. 

To address these limitations, in this paper, we propose a unified \underline{He}te\underline{ro}geneous \underline{Con}trastive Learning framework, named \method, which jointly models the view and label heterogeneity using two contrastive loss terms. 
In particular, to leverage the large amount of unlabeled data, we propose a weighted unsupervised contrastive loss to alleviate the potential negative impact of false negative pairs, which automatically adjusts the weights of the samples drawn from the negative set; to leverage the limited labeled data, we propose a weighted supervised contrastive loss to group the samples with similar label vectors together in the latent space, where the weights reflect how similar the label vectors of two samples are. By combining these two contrastive loss terms, our proposed framework is capable of modeling multiple types of heterogeneity in the presence of limited label information. We provide the theoretical analysis showing that the vanilla contrastive learning loss may easily lead to the sub-optimal solution in case of false negative pairs, whereas the proposed weighted loss could automatically adjust the weight based on the similarity of the learned representations to mitigate this issue. In addition, we show that our proposed weighted unsupervised loss is the lower bound of the mutual information between the hidden representation of two views of the same sample and that the weighted supervised contrastive loss is the lower bound of the mutual information between two samples sharing similar label information.
Our main contributions are summarized below:
\begin{itemize}
    \item A novel framework for deep heterogeneous contrastive learning, which effectively leverages large amount of unlabeled data in the presence of limited labeled information.
    \item Theoretical analysis to show that the vanilla contrastive learning loss easily leads to the sub-optimal solution and that the two weighted contrastive losses are two lower bounds of the mutual information.
    \item Experimental results on real-world data sets demonstrating the effectiveness and efficiency of the proposed framework.
\end{itemize}
The rest of this paper is organized as follows. After a brief review of the related work in Section 2, we introduce our proposed framework for heterogeneous contrastive learning in Section 3.  The systematic evaluation of the proposed framework on real-world data sets is presented in Section 4 before we conclude the paper in Section 5.

%% file: Related_Work.tex
\section{Related Work}
In this section, we briefly review the related work on contrastive learning and heterogeneous learning.

\subsection{Contrastive Learning}
Recently, self-supervised learning~\cite{wu2021indirect,song2020multi, chuang2020debiased, khosla2020supervised, tian2019contrastive, chen2020simple, zheng2021tackling, jing2021hdmi,li2022graph,feng2022adversarial, jing2021graph} attracts researchers' great attention due to its prominent performance modeling the unlabeled data. ~\cite{oord2018representation} is one of the earliest works in contrastive learning, which proposes the contrastive predictive coding framework (Info-NCE) to extract useful information from high dimensional data with a theoretical guarantee. Based on this work, recent studies reveal a surge of research interest in contrastive learning. 
~\cite{khosla2020supervised} extends Info-NCE to the supervised scenario and consider the situation where the hidden representation of the samples from the same class should be close to each other in the latent space.  ~\cite{chen2020simple} proposes a simple framework for contrastive learning of visual representations, which boosts the performance of supervised and semi-supervised tasks on Image-Net.  ~\cite{tian2019contrastive} extends Info-NCE to a multi-view setting and learns a representation to maximize the mutual information between different views of the same sample. However, ~\cite{tian2019contrastive} fails to consider the potential similarity of the hidden representations between the positive sample and the sample drawn from negative sets. Similarly, supervised contrastive loss (SupCon)~\cite{khosla2020supervised} could not be directly applied to handle the multi-label scenario. The authors of ~\cite{huo2020heterogeneous} propose a novel contrastive learning method to alleviate the mismatch between the contrastive objective and data augmentation operations by adding spatial information. In this paper, we propose both the weighted unsupervised contrastive loss and the weighted supervised contrastive loss to maximize the mutual information between the hidden representation of two views from the same sample and the mutual information between the hidden representations of two samples from the same class. 

\subsection{Heterogeneous Learning}
In this subsection, we briefly review the recent works on multi-view learning, multi-label learning, and multi-class learning. 
Multi-view learning\cite{DBLP:journals/pami/XuTX15,DBLP:journals/tip/XuT015,DBLP:conf/cikm/FuXLTH20, zhou2015muvir} has been studied for decades and the initial works mainly focus on co-training~\cite{DBLP:conf/colt/BlumM98}, multiple kernel method~\cite{DBLP:conf/icml/LanckrietCBGJ02}, and subspace learning~\cite{DBLP:journals/corr/abs-cs-0609071}.
Recently, more and more attention is paid to the direction of subspace learning. ~\cite{DBLP:journals/pami/XuTX15} proposes a multi-view intact space learning method by integrating the encoded complementary information from multiple views to discover a latent representation;
~\cite{DBLP:journals/tip/NieCLL18} performs semi-supervised classification and local structure learning simultaneously, and automatically allocates weight for each view.
In multi-label learning, ~\cite{DBLP:journals/tkde/ZhuKZ18} proposes to exploit both global and local label correlations to learn a latent representation for both full-label and missing label scenarios.
~\cite{DBLP:conf/kdd/XuT016} tackles the extreme multi-label scenario and proposes a novel low-rank matrix decomposition method to handle the long tail problem with theoretical analysis. ~\cite{DBLP:journals/ijon/PupoMV15} aims to transform the multi-label problem into a single-label problem, and a distance function is defined to reduce the negative impact of the noisy features.
~\cite{DBLP:conf/aaai/HuangGZ14} constructs a low-dimensional subspace shared by all labels and exploiting label relations within the shared subspace.
In multi-class learning, ~\cite{DBLP:conf/nips/Sohn16} proposes a novel metric learning objective function called multi-class N-pair loss allowing joint comparison among N-1 negative examples. ~\cite{DBLP:journals/ijcv/YangMNCH15} presents a semi-supervised multi-class active learning method by exploiting the active pool to evaluate the uncertainty of data and imposing a diversity constraint to select the diverse data. Different from these methods, we propose the weighted unsupervised contrastive loss to maximize the mutual information between two views, and model the label correlation by maximizing the mutual information between the samples with the same label via weighted supervised contrastive loss.

%% file: Algorithm.tex
\section{Proposed \method\ Framework}
In this section, we introduce our proposed framework for heterogeneous contrastive learning named \method. We start by introducing the notation and then present the overall loss function with two regularization terms leveraging the labeled and unlabeled data with the theoretical analysis, respectively.

\subsection{Notation}
Throughout this paper, we use lower-case letters for scalars (e.g., $\gamma$), 
and a bold upper-case letter for a matrix (e.g., $\bm{X}$). We assume that the input data $\mathcal{D}$ consists of two parts, namely $\mathcal{D} = \{\mathcal{L}, \mathcal{U}\}$.
We use  $\mathcal{L} =\{\bm{X^\mathcal{L}}, \bm{Y^\mathcal{L}}\}$ to denote the labeled data set, where $\bm{X^\mathcal{L}} \in \mathbb{R}^{n \times d}$ and $\bm{Y^\mathcal{L}}\in \mathbb{R}^{n \times c}$  are the input feature and binary label matrices for the labeled data set, respectively. $n$ is the number of the labeled samples, $d$ is the dimensionality of the input features, and $c$ is the number of labels. $\bm{Y^\mathcal{L}}_i(a)$ is the $a^{th}$ binary label of sample $\bm{X_i^\mathcal{L}}$.
Similarly, we denote $\mathcal{U} =\{\bm{X^\mathcal{U}}\}$ as the unlabeled data set, where $\bm{X^\mathcal{U}} \in \mathbb{R}^{m \times d}$ is the input feature matrix for the unlabeled data set and $m$ is the number of the unlabeled samples.
Let $\bm{Z^\mathcal{L}}$ and $\bm{Z^\mathcal{U}}$ be the hidden representations of labeled data and unlabeled data generated by the encoder $\bm{E}(\cdot)$ respectively, \ie, $\bm{Z^\mathcal{L}}=\bm{E}(\bm{X^\mathcal{L}})$, $\bm{Z^\mathcal{U}}=\bm{E}(\bm{X^\mathcal{U}})$.
For the ease of explanation, we denote $\bm{X_i}$ as a sample from either the labeled data set or the unlabeled data set when there is no confusion in a specific context and $\bm{Z_i}$ as the hidden representation of $\bm{X_i}$.
Furthermore, in the presence of view heterogeneity, we assume that sample $\bm{X_i}$ is characterized by two views although the proposed techniques can be readily extended by multiple views
~\footnote{If only one view is available, we could use two different data augmentation methods to generate two views by, e.g., following the strategy mentioned in~\cite{chen2020simple}.}
: we denote $\bm{X_{i,1}}$ and $\bm{X_{i,2}}$ as the first and second views of $\bm{X_i}$, respectively. For the two views, we could use two different encoders $\bm{E_1}$ and $\bm{E_2}$ to obtain the corresponding hidden representation $\bm{Z_{i,1}}$ and $\bm{Z_{i,2}}$, where $\bm{Z_{i,1}}=\bm{E_1}(\bm{X_{i,1}})$ and $\bm{Z_{i,2}}=\bm{E_2}(\bm{X_{i,2}})$ are the representation extracted from the first view and the second view, respectively.

\subsection{Objective Function}
Now, we are ready to introduce the overall objective function:
\begin{equation}
    \label{overall}
    \begin{split}
        \min J &= L_c(\bm{Y^\mathcal{L}}, \bm{\hat{Y}^\mathcal{L}})  + \alpha L_u(\bm{X^\mathcal{L}}, \bm{X^\mathcal{U}}, \bm{Z^\mathcal{L}}, \bm{Z^\mathcal{U}}) \\
        & + \beta L_s(\bm{Z^\mathcal{L}}, \bm{Y^\mathcal{L}})
    \end{split}
\end{equation}
where $\hat{Y}^\mathcal{L}\in  \mathbb{R}^{n \times c}$ is the prediction made by the classifier $\bm{C}(\cdot)$, \ie, $\hat{Y}^\mathcal{L}=\bm{C}(\bm{Z^\mathcal{L}})$, $L_c$ is the cross entropy loss, $L_u$ is the unsupervised contrastive loss to model multi-view heterogeneity by regularizing the hidden feature representations $\bm{Z^\mathcal{L}}$ and $\bm{Z^\mathcal{U}}$, $L_s$ is the supervised contrastive loss to model multi-label or multi-class heterogeneity by regularizing the hidden feature representation $\bm{Z^\mathcal{L}}$, and $\alpha$ and $\beta$ are two positive hyper-parameters balancing the two regularization terms. Next, we elaborate on each regularization term respectively.

\subsubsection{$L_u$: Weighted Unsupervised Contrastive Loss}
The main idea of the unsupervised contrastive loss is to utilize the rich unlabeled data to enhance the quality of the hidden representation. Following ~\cite{song2020multi}, it can be written as follows:

\begin{equation}
    \label{lu0}
    \begin{split}
         L = -\E_{X_i\in \mathcal{D}}[ \log \frac{f(\bm{X_i}, \bm{Z_i})}{f(\bm{X_i}, \bm{Z_i}) + \sum_{k\neq i} f(\bm{X_i}, \bm{Z_k})}]
    \end{split}
\end{equation}
where $f(\cdot,\cdot)$ is the similarity measurement function, \eg, $f(\bm{X_i}, \bm{Z_i})=\exp(\frac{\bm{X_i}^T \bm{Z_i}}{\tau})$, where $\tau$ is the temperature, and $\bm{X_i}$ is a sample drawn from $\mathcal{D}$. Following ~\cite{song2020multi}, $(\bm{X_i}, \bm{Z_i})$ in the numerator is considered as a positive pair and $(\bm{X_i}, \bm{Z_k})$ in the denominator is considered as a negative pair. Eq.~\ref{lu0} aims to maximize the mutual information between the original input features and the hidden representations by minimizing the unsupervised contrastive learning loss. Similarly, ~\cite{chen2020simple} proposes to maximize the similarity between two augmented views of the same sample denoted as $\bm{X_{i,1}}$ and $\bm{X_{i,2}}$, which can be formulated as follows: 

\begin{equation}
    \label{lu1}
    \begin{split}
         L_1 = -\E_{X_i\in \mathcal{D}}[ \log \frac{f(\bm{Z_{i,1}}, \bm{Z_{i,2}})}{f(\bm{Z_{i,1}}, \bm{Z_{i,2}}) + \sum_{k\neq i} f(\bm{Z_{i,1}}, \bm{Z_{k,2}})}]
    \end{split}
\end{equation}

Though these two unsupervised contrastive loss functions take advantage of the rich information from the unlabeled data, neither of them take into consideration the scenario where two samples with similar input features tend to have similar hidden representations. Basically, we could consider sample B in Figure~\ref{fig_motivation} as a false-negative sample for sample A as they share the same label information, and sample C as a true negative sample for sample A due to the different label information (\eg, cat vs dog). 
Formally, we could define the false negative sample and the true negative sample as follows:

\begin{definition}
    \label{definition_2}
    Given an unlabeled sample $\bm{X_i}$, we say sample $\bm{X_j}$ is a false negative sample of $\bm{X_i}$, if their optimal representations satisfy $e^{(\bm{Z_i}^*)^T\bm{Z_j}^*/\tau} > 1$ for some small positive value $\tau$. Similarly, we say sample $\bm{X_k}$ is a true negative sample of $\bm{X_i}$, if their optimal representations satisfy $e^{(\bm{Z_i}^*)^T\bm{Z_k}^*/\tau} \approx 0$ for some small positive value $\tau$.
\end{definition}

\begin{lemma}
\label{lemma_u1}
Given the vanilla contrastive learning loss function $L_1$, if there exists one false negative sample in the batch during training, the contrastive learning loss will lead to a sub-optimal solution.
\end{lemma}
\textbf{Proof:} In Appendix.

Lemma~\ref{lemma_u1} shows that the vanilla contrastive learning loss will easily lead to a sub-optimal solution with only one false negative sample. A naive way to alleviate this problem is to re-weight all negative pairs based on the similarity of two original input features as follows:
\begin{equation}
    \nonumber
    \begin{split}
          L_2=-\E_{X_i\in \mathcal{D}}[ \log \frac{f(\bm{X_i}, \bm{Z_i})}{f(\bm{X_i}, \bm{Z_i}) + \sum_{\bm{X_{k}}\in \mathcal{N}_{i}^\mathcal{D}} sim(\bm{X_i}, \bm{X_k})f(\bm{X_i}, \bm{Z_k})}]
    \end{split}
\end{equation}
where $sim(\bm{X_i}, \bm{X_k})$ is a similarity measurement between $\bm{X_i}$ and $\bm{X_k}$, \eg, $sim(\bm{X_i}, \bm{X_k})=exp(1-\frac{\bm{X_i} \cdot\bm{X_k}}{|\bm{X_i}| |\cdot\bm{X_k}|})$, and $\mathcal{N}_{i}^\mathcal{D} =\mathcal{D} \backslash \{i\}$ is the negative set consisting of the entire data set except for $\bm{X_i}$.  The intuition of this equation is that if two samples chosen as a negative pair are similar in terms of the input feature similarity, they are very likely to have the similar hidden representations. Thus, we reduce the weight of this negative pair based on how similar their original features are. However, the computational cost for $sim(\bm{X_i}, \bm{X_k})$ is extremely expensive for high dimensional data, such as images. To reduce the computational costs, we propose a novel weighted unsupervised contrastive learning loss to re-weight the negative pairs based on the projected low dimensional representations instead of the original input features as follows:

\begin{align}
    \label{lu2}
    \scalebox{0.98}{
         $L_u = -\E_{X_i\in \mathcal{D}}[ \log \frac{f(\bm{X_i}, \bm{Z_i})}{f(\bm{X_i}, \bm{Z_i}) + \sum_{\bm{X_{k}}\in \mathcal{N}_{i}^\mathcal{D}} g(\bm{Z_i}, \bm{Z_k})f(\bm{X_i}, \bm{Z_k})}]$ } \\ 
    \nonumber \scalebox{0.98}{     $g(\bm{Z_i}, \bm{Z_k}) = \frac{1}{2}(exp(1-\frac{\bm{Z_i}^T\bm{H}(\bm{Z_k})}{|\bm{Z_i}||\bm{H}(\bm{Z_k})|})
         + exp(1-\frac{\bm{Z_k}^T\bm{H}(\bm{Z_i})}{|\bm{Z_k}||\bm{H}(\bm{Z_i})|}))$}
\end{align}

where $\bm{H}(\cdot)$ is a fully-connected layer followed by an activation function (\eg, the sigmoid function). The intuition of this design is that if two samples chosen as a negative pair have the similar hidden representations, we aim to reduce the weight of this negative pair by the weighting function $g(\bm{Z_i}, \bm{Z_k})$. In other word, if $\bm{Z_i}$ and $\bm{Z_k}$ are dissimilar, then the value of the weighting function $g(\bm{Z_i}, \bm{Z_k})$ is expected to be large and minimizing $L_u$ will further push $\bm{Z_i}$ away from $\bm{Z_k}$. If $\bm{Z_i}$ and $\bm{Z_k}$ are similar, the value of $g(\bm{Z_i}, \bm{Z_k})$ is expected to be small and minimizing $L_u$ will reduce the weight between $\bm{Z_i}$ and $\bm{H}(\bm{Z_k})$ instead of pushing $\bm{Z_i}$ away from $\bm{Z_k}$.

Similarly, the proposed weighted unsupervised contrastive loss can be naturally extended to model multi-view data. Different from the intuition of contrastive learning for a single view, the multi-view contrastive loss aims to maximize the mutual information of the hidden representations of two views.
More specifically, given a sample $\bm{X_i}$ with two views $\bm{X_{i,1}}$ and $\bm{X_{i,2}}$, the weighted unsupervised contrastive loss could be updated as follows:
\begin{equation}
    \label{lu3}
    \scalebox{0.99}{
         $L_u  = -\E_{X_i\in \mathcal{D}}[  \log \frac{f(\bm{Z_{i,1}}, \bm{Z_{i,2}})}{f(\bm{Z_{i,1}}, \bm{Z_{i,2}}) + \sum_{\bm{X_{k}} \in \mathcal{N}_{i}^\mathcal{D}} g(\bm{Z_{i,1}}, \bm{Z_{k,j}}) f(\bm{Z_{i,1}}, \bm{Z_{k, j}})} ]$}
\end{equation}
where we denote $\bm{X_{k,j}}$ to be the $j^{\textrm{th}}$ view of $\bm{X_k}$, $\bm{Z_{k,j}}$ is the hidden representation of $\bm{X_{k,j}}$ and $\mathcal{N}_{i}^\mathcal{D} =\mathcal{D} \backslash \{i\}$. 
This equation aims to maximize the mutual information between the hidden representations extracted from two views, and to minimize the similarity of the hidden representations extracted from two different samples. Notice that in the denominator of this equation, we follow~\cite{chen2020simple} to include both the first view and the second view of $\bm{X_k}$ as the negative samples in order to increase the size of the negative set. As the size of the negative set increases, we tend to have a tighter lower bound, which is demonstrated in Lemma~\ref{lemma_u} and Section 4.5 Parameter Analysis. The extension to more than two views is straightforward
, and we omit it for brevity.

\begin{table}
\centering
\begin{tabular}{|*{3}{c|}}
\hline \textbf{-}       & \textbf{Noisy MNIST} & \textbf{CelebA}  \\
\hline Setting   & Multi-class  &   Multi-label  \\
\hline Number of labels     &     10  & 40  \\
\hline Size of data set     &     70,000  & 202,599  \\
\hline Number of unique label vectors     &     10  & 115,114  \\
\hline Average size of positive set  & 7,000 & 1.76 \\
\hline
\end{tabular}
\caption{Statistics of label information for two data sets}
\vspace{-0.5cm}
\label{table1}
\end{table} 

\subsubsection{$L_s$: Weighted Supervised Contrastive Loss}
The goal of the supervised contrastive loss is to maximize the mutual information between two samples with the same label~\cite{khosla2020supervised}. In the binary classification setting (the number of binary labels $c=1$), we denote the set of positive samples drawn from the labeled data set as $\mathcal{P}^\mathcal{L}=\{\bm{X_j}|Y_j^\mathcal{L}=1\}$ and the set of negative samples drawn from the labeled data set as $\mathcal{N}^\mathcal{L}=\{\bm{X_k}|Y_k^\mathcal{L} \neq 1\}$. Based on ~\cite{khosla2020supervised}, the supervised contrastive learning loss (SupCon) is formulated as follows:
\begin{equation}
    \label{ls1}
    \begin{split}
         L_{sup} = -\E_{\bm{X_i}, \bm{X_j} \in \mathcal{P}^\mathcal{L}} [ \log \frac{f(\bm{S_i}, \bm{S_j})}{f(\bm{S_i}, \bm{S_j}) + \sum_{\bm{X_k} \in \mathcal{N}^\mathcal{L}} f(\bm{S_i}, \bm{S_k})} ]
    \end{split}
\end{equation}
where $\bm{S_i}=concat(Z_{i,1}, Z_{i,2})$ is the concatenation of the hidden representations for the two views of $\bm{X_i}$ ($\bm{S_i}=Z_{i,1}$ if only one view is available). The intuition of this equation is that any pair of samples drawn from the positive set $\mathcal{P}^\mathcal{L}$ should be closer than the samples from the negative set $\mathcal{N}^\mathcal{L}$ in the latent space.
Despite its superior performance, SupCon is not designed for the multi-label setting. Different from the binary classification problem or multi-class problem where a sample could only be classified into one class, in the multi-label setting, a sample could be characterized with multiple labels. As the number of the labels $c$ increases, it becomes harder to find two samples with the same label vector (as there are $2^c$ different combinations for $c$ different binary labels).
For example, Table ~\ref{table1} shows the statistics of label information for the Noisy MNIST data set~\cite{WangALB15} and the CelebA data set~\cite{liu2015deep}. By observation, we could see that in the multi-label setting, there are 115,114 unique label vectors on the CelebA data set, and the average size of the positive set is only 1.76, which is largely different from that for the Noisy MNIST data set in the multi-class setting. This indicates that SupCon is not applicable in the multi-label setting as it is impossible to construct the positive set that contains at least two samples for each unique label vector for contrastive learning.

To overcome this issue, we propose the weighted supervised contrastive loss formulated as follows:

\begin{align}
    \label{ls2}
    \scalebox{0.95}{ $L_s = -\frac{1}{c}\sum_{a=1}^c \E_{\bm{X_i},\bm{X_j} \in \mathcal{P}^\mathcal{L}(a)} [ \log \frac{\sigma f(\bm{S_i}, \bm{S_j})}{\sigma f(\bm{S_i}, \bm{S_j}) + \sum_{\bm{X_k} \in \mathcal{N}^\mathcal{L}(a)}  \gamma f(\bm{S_i}, \bm{S_k})} ]$ } \\
    \nonumber\scalebox{0.97}{$\sigma = 1- dist(\bm{Y_i}^\mathcal{L}, \bm{Y_j}^\mathcal{L})/c, ~~\gamma= dist(\bm{Y_i^\mathcal{L}}, \bm{Y_k}^\mathcal{L})$ }
\end{align}
where $dist(\bm{Y_i}^\mathcal{L}, \bm{Y_k}^\mathcal{L})$ is the distance measurement between two label vectors, \eg, the hamming distance, $\mathcal{P}^\mathcal{L}(a)=\{\bm{X_j}|Y_j^\mathcal{L}(a)=1\}$ is the set of positive samples drawn from the labeled data set in terms of the $a^{th}$ label and $\mathcal{N}^\mathcal{L}(a)=\{\bm{X_k}|Y_k^\mathcal{L}(a) \neq 1\}$ is the set of negative samples.
The intuition of Eq.~\ref{ls2} is that the samples with similar label vectors should be close to each other in the latent space, and the magnitude of the similarity is determined based on how similar their label vectors are. Specifically, in the numerator of Eq.~\ref{ls2}, we aim to maximize the similarity between the hidden representations of $\bm{X_i}$ and $\bm{X_j}$ if the $a^{th}$ binary label for these two samples are both positive, \ie, $\bm{Y_i}^\mathcal{L}(a) = \bm{Y_j}^\mathcal{L}(a)=1$ in the multi-class setting. However, in the multi-label setting, since one sample could be characterized by multiple labels, we reweight the similarity of the hidden representations by the function $\sigma$ such that if the label vectors of the two samples are identical, $\sigma$ is equal to 1, and it gradually approaches 0 as the two label vectors become completely different. Similarly, in the denominator, we aim to minimize the similarity between the hidden representations of samples $\bm{X_i}$ and $\bm{X_k}$ if their $a^{th}$ labels are different and the similarity measurement is also weighted based on how dissimilar their label vectors are. 

\subsection{Special Cases}
The existing contrastive losses proposed in SupCon~\cite{khosla2020supervised} and SimCLR~\cite{chen2020simple} can be considered as special cases of our proposed framework.
First of all, the weighted supervised contrastive loss in our proposed method can be degraded to SupCon. In the binary classification setting or multi-class setting, the distance measurement function $dist(\bm{Y_i}^\mathcal{L}, \bm{Y_j}^\mathcal{L})$ in Eq.~\ref{ls2} can be reduced to an indicator function $dist(\bm{Y_i}^\mathcal{L}, \bm{Y_j}^\mathcal{L})=\bm{1}_{Y_i^\mathcal{L} \neq Y_j^\mathcal{L}}$, where $\bm{1}_{Y_i^\mathcal{L} \neq Y_j^\mathcal{L}}=0$ if $\bm{Y_i}^\mathcal{L}=\bm{Y_j}^\mathcal{L}$ and $\bm{1}_{Y_i^\mathcal{L} \neq Y_j^\mathcal{L}}=1$ otherwise (as $\bm{Y_i}^\mathcal{L}$ and $\bm{Y_j}^\mathcal{L}$ are scalars in the binary classification setting or multi-class setting).
In this case, the weight imposed on the positive pair in the numerator of Eq.~\ref{ls2} is reduced to $\sigma = 1- dist(\bm{Y_i}^\mathcal{L}, \bm{Y_j}^\mathcal{L})/c=1-\bm{1}_{\bm{Y_i}^\mathcal{L} \neq \bm{Y_j}^\mathcal{L}}=1$ because $\bm{Y_i}^\mathcal{L} = \bm{Y_j}^\mathcal{L}$ for any positive pairs.
Similarly, the weight $dist(\bm{Y_i}^\mathcal{L}, \bm{Y_k}^\mathcal{L})=\bm{1}_{\bm{Y_i}^\mathcal{L} \neq \bm{Y_k}^\mathcal{L}}$ imposed on the denominator is equal to 1 because $\bm{Y_i}^\mathcal{L} \neq \bm{Y_k}^\mathcal{L}$ for any negative pairs. Thus, in the binary classification or multi-class setting, Eq.~\ref{ls2} could be reduced to Eq.~\ref{ls1}, which is exactly the formulation of SupCon. Compared with SupCon, our proposed method can not only handle the multi-class problem but also the multi-label classification problem.
Similarly, the weighted unsupervised contrastive loss in our proposed method becomes the objective function in SimCLR by setting the weights of all negative pairs to 1.

\subsection{Theoretical Analysis}
In this subsection, we provide the analysis regarding the properties of the two proposed contrastive losses.

\begin{lemma}
\label{lemma_s}
Given two samples $\bm{X_i}$ and $\bm{X_j}$ from the same class drawn from the labeled set $\mathcal{L}$, we have $I(\bm{X_i}, \bm{X_j}) \geq -\frac{1}{\sigma} (L_s - N)$, where $I(\bm{X_i}, \bm{X_j})$ is the mutual information between $\bm{X_i}$ and $\bm{X_j}$,  $L_s$ is the supervised contrastive loss weighted by hamming distance measurement, 
$\sigma = 1- dist(\bm{Y_i}^\mathcal{L}, \bm{Y_j}^\mathcal{L})/c$,  which measures the ratio of two binary labels for two samples $\bm{X_i}$ and $\bm{X_j}$ having the same value,
and $N=\frac{1}{c}\sum_{a=1}^c \log(|\mathcal{N}^\mathcal{L}(a)|)$.
\end{lemma}
\textbf{Proof:} In Appendix.
\begin{lemma}
\label{lemma_u}
Given a sample $\bm{X_i}$ drawn from the entire set $\mathcal{D}$, we have $I(\bm{X_{i,1}}, \bm{X_{i,2}}) \geq - L_u + \log(|\mathcal{N}^\mathcal{D}_i|)$, where $I(\bm{X_{i,1}}, \bm{X_{i,2}})$ is the mutual information between $\bm{X_{i,1}}$ and $\bm{X_{i,2}}$, $L_u$ is the unsupervised contrastive loss weighted by $g(\bm{Z_{i,1}}, \bm{Z_{k,j}})$ and $|\mathcal{N}^\mathcal{D}_i|$ is the size of the negative set.
\end{lemma}
\textbf{Proof:} In Appendix.

Based on the Lemma~\ref{lemma_s}, we observe that the proposed weighted supervised contrastive loss is the lower bound of the mutual information of two samples sharing similar label information. In Lemma~\ref{lemma_u}, we prove that the weighted unsupervised contrastive loss is the lower bound of the mutual information between the hidden representations of two views of the same sample. As the size of the data set becomes larger, the lower bound becomes tighter, which is further demonstrated in Subsection 4.5 Parameter Analysis. Combining both weighted supervised contrastive loss and weighted unsupervised contrastive loss, we aim to explore the hidden representations that enjoy the following benefits: (1) if two samples are from the same class, then their hidden representations should be close to each other in the embedding space by minimizing $L_s$; (2) the hidden representations should only contain the information shared by the two views and discard the irrelevant information as much as possible by minimizing $L_u$.

%% file: Experiment.tex
\section{Experimental Results}
In this section, we demonstrate the performance of our proposed framework in terms of effectiveness by comparing it with state-of-the-art methods. In addition, we conduct a case study to show how different levels of noise influence our proposed methods, which is followed by the parameter analysis and efficiency analysis (in appendix A.1).

\subsection{Experiment Setup}
\textbf{Data Sets:}
We mainly evaluate our proposed algorithm on the following data sets: Noisy MNIST (N-MNIST) ~\footnote{\url{http://yann.lecun.com/exdb/mnist/}}; X-ray Microbeam (XRMB) ~\footnote{\url{https://ttic.uchicago.edu/~klivescu/XRMB_data/full/README}}, Celebrity Face Attributes (CelebA) ~\footnote{\url{http://mmlab.ie.cuhk.edu.hk/projects/CelebA.html}} and Scene~\footnote{\url{http://mulan.sourceforge.net/datasets-mlc.html}}. 
N-MNIST~\cite{basu2017learning} data set consists of 70,000 images of handwritten digits with an additive white Gaussian noise added to the MNIST data set. Specifically, we add Gaussian noise to MNIST~\cite{lecun1998gradient} data set to generate the N-MNIST data set by following the strategy introduced in~\cite{WangALB15}. We first rescale the pixel values of each image to [0,1], then add the random noise uniformly sampled from [0, 1] to each pixel, and finally truncate the pixel values to [0, 1].
Scene~\cite{boutell2004learning} is a single-view multi-label data set characterized with six binary labels, which consists of 2,407 samples. 
XRMB~\cite{westbury1994x} is a multi-view multi-class data set, which consists of 40 binary labels and two views. The first view is acoustic data with 273 features and the second view is articulatory data with 112 features. 
CelebA~\cite{liu2015deep} is a large-scale face attributes data set with more than 200K celebrity images, labeled with 40 attributes. Following the strategy used in~\cite{chen2020simple}, we use two data augmentation methods, \eg, (1) crop and resize and (2) color distortion, to generate two views for the CelebA data set.\\
\begin{table*}
\centering
\caption{Results on Scene and N-MNIST data sets. Notice that \method-s is identical to SupCon in multi-class setting (\eg, in N-MNIST data set).}
\begin{tabular}{|*{5}{c|}}
\hline - & \multicolumn{2}{c|}{Scene} & \multicolumn{2}{c|}{N-MNIST}\\
\hline \textbf{Model}       & \textbf{F1 Score}     & \textbf{AUC}          & \textbf{F1 Score}     & \textbf{AUC} \\
\hline DNN                  & 0.5902 $\pm$ 0.0122   & 0.8647 $\pm$ 0.0077   & 0.9036 $\pm$ 0.0045   & 0.9463 $\pm$ 0.0025 \\   
\hline Info-NCE             & 0.6144 $\pm$ 0.0072   & 0.8783 $\pm$ 0.0110   & 0.8877 $\pm$ 0.0064   & 0.8900 $\pm$ 0.0067 \\
\hline SupCon               & 0.5969 $\pm$ 0.0207   & 0.8624 $\pm$ 0.0148   & 0.9265 $\pm$ 0.0040   & 0.9584 $\pm$ 0.0037 \\
\hline MIB                  & 0.6151 $\pm$ 0.0147   & 0.8759 $\pm$ 0.0114   & 0.8920 $\pm$ 0.0117   & 0.8051 $\pm$ 0.0086 \\
\hline C2AE                 & 0.6145 $\pm$ 0.0275   & 0.8834 $\pm$ 0.0105   & 0.9049 $\pm$ 0.0104   & 0.9120 $\pm$ 0.0099 \\
\hline DeepMTMV             & 0.6173 $\pm$ 0.0200   & 0.8727 $\pm$ 0.0113   & 0.9051 $\pm$ 0.0043   & 0.9474 $\pm$ 0.0023 \\
\hline \method-s            & 0.6104 $\pm$ 0.0166   & 0.8807 $\pm$ 0.0117   & 0.9265 $\pm$ 0.0040   & 0.9584 $\pm$ 0.0037 \\
\hline \method-u            & 0.6231 $\pm$ 0.0181   & 0.8831 $\pm$ 0.0118   & 0.9246 $\pm$ 0.0028   & 0.9582 $\pm$ 0.0016 \\
\hline \method              & \textbf{0.6366 $\pm$ 0.0141} & \textbf{0.8878 $\pm$ 0.0079} & \textbf{0.9363 $\pm$ 0.0019} & \textbf{0.9705 $\pm$ 0.0012} \\
\hline
\end{tabular}
\label{table_single_view}
\end{table*} 
\noindent\textbf{Experiment Setting:}
The neural network structure of the proposed method is manually adjusted based on the input data type. The neural network structure and two hyper-parameters $\alpha$ and $\beta$ for each data set will be specified in Subsection \ref{single_view_multi_label} and \ref{multi_view_multi_label}. For each data set, we randomly draw the same number of training samples, repeat the experiments 5 times, and report the mean and the standard deviation of the F1 score and AUC value. In all experiments, we set the initial learning rate to be 0.05 and the optimizer is momentum stochastic gradient descent with Layer-wise Adaptive Rate Scaling scheduler (LARS)~\cite{you2017large}. Besides, we consider the test set as the unlabeled set $\mathcal{U}$, the similarity function $f(a,b)$ is defined as  $f(a, b)=\exp(\frac{a \cdot b}{|a||b|})$ and $dist(\bm{Y_i}^\mathcal{L}, \bm{Y_k}^\mathcal{L})$ is the hamming distance measurement. \\
\noindent \textbf{Reproducibility:}
All of the real-world data sets are publicly available. The code of our algorithms could be found in the link~\footnote{\url{https://github.com/Leo02016/HeroCon}}. The experiments are performed on a Windows machine with a 16GB RTX 5000 GPU.\\
\noindent \textbf{Comparison Methods:}
In our experiments, we compare our proposed method, \ie, ~\method~ with the following methods: 
\begin{itemize}
    \item DNN: a simple deep neural network, the structure of which will be specified for each data set; 
    \item MIB~\cite{federici2020learning}: a multi-view information bottleneck based method that only retains the information relevant to the labels and minimizes the use of other information;
    \item CMC~\cite{tian2019contrastive}: a contrastive multi-view coding method that maximizes the similarity of multiple views; 
    \item Info-NCE~\cite{oord2018representation}: a contrastive learning method for self-supervised learning in the single view setting; 
    \item SupCon~\cite{khosla2020supervised}: a supervised contrastive learning method in the single view setting; 
    \item DeepMTMV~\cite{DBLP:conf/sdm/ZhengCH19}: a deep framework modeling both view heterogeneity and label heterogeneity; 
    \item C2AE~\cite{yeh2017learning}: a canonical correlated auto-encoder based method for multi-label classification problem, which proposes a label-correlation sensitive loss function to exploit label dependency;
    \item \method-\textit{u}: the first variant of our proposed method by discarding the weighted supervised contrastive loss term; 
    \item \method-\textit{s}: the second variant of our proposed method by discarding the weighted unsupervised contrastive loss term.
\end{itemize}
The neural network architecture of CMC, Info-NCE, SupCon, MIB, and our methods will be specified for each data set in different experimental settings. As some proposed methods are only designed for a particular setting, we only report the performance of these baselines if applicable. 
For data sets in the multi-view setting, we concatenate the hidden representations for SupCon and C2AE.\\
\textbf{Efficiency Analysis:} Due to the space limit, we move the efficiency to the Appendix and it could be found in A.1.

\subsection{Single-view Multi-label Setting}
\label{single_view_multi_label}
In this subsection, we test the performance of our proposed method on two real-world data sets in the single-view multi-label setting, including Scene data set and N-MNIST data set. In the experiments, we use Eq.~\ref{lu2} to compute the weighted unsupervised contrastive loss for ~\method~ and ~\method-u.

For the Scene data set, we randomly sample $5\%$ data (120 samples) as the training set and the remaining $95\%$ data as the test set. The number of binary labels is 6. The neural network architectures of DNN, Info-NCE, SupCon, and our methods are the same, which is a three-layer fully-connected neural network. Two hyper-parameters $\alpha$ and $\beta$ for ~\method~ are $0.7$ and $0.02$, respectively; the hyper-parameter $\alpha$ for ~\method-${u}$ is $0.3$; and the hyper-parameter $\beta$ for ~\method-${s}$ is $0.01$. The batch size is the entire training set, the number of epochs for our methods is 200 and the size of the negative set $|\mathcal{N}_i|$ is equal to 2,406 ($|\mathcal{D}|$ - 1). Table~\ref{table_single_view} shows the performance of our proposed methods and state-of-the-art methods. By observation, we find that our proposed method~\method ~and \method-u outperform all baseline models. Specifically, compared with Info-NCE, \method-u and \method~ further boost the performance by 0.9\% and 2.2\% in terms of F1 score, respectively. This suggests that Info-NCE indeed leads to sub-optimal performance as it assigns the same weight to each negative sample no matter how similar this negative sample is to the positive sample. From this table, we also observe that DeepMTMV, MIB and C2AE have similar performance but our proposed method improves the performance by more than 1.9\%.  Our conjecture is that \method~ takes advantage of the rich unlabeled data, and it is capable of learning better representations in the case of limited labeled samples.

For the N-MNIST data set, we sampled 20 images for each digit from 10,000 images as our training set and the remaining 60,000 samples are considered as our test set.
The neural network architectures of DNN, Info-NCE, SupCon, and our methods are the same, which is a two-layer convolutional layer followed by a max-pooling layer and a two-layer fully-connected neural network. Two hyper-parameters $\alpha$ and $\beta$ for ~\method~ are $0.1$ and $1$, respectively; the hyper-parameter $\alpha$ for ~\method-${u}$ is $0.1$; and the hyper-parameter $\beta$ for ~\method-${s}$ is $2$. The batch size is 200 (the size of the entire labeled set), the number of iterations for our methods is 500 and the size of the negative set $|\mathcal{N}_i|$ is 4,199 (200 labeled samples and 4000 unlabeled samples for each iteration). Based on Table~\ref{table_single_view}, we observe that \method~ achieves the best F1 score and AUC (as we mentioned in Section 3.3, in multi-class setting, \method-s would degrade to SupCon and thus their performance is the same).
Different from the performance improvement in the Scene data set for unsupervised contrastive learning methods, in the N-MNIST data set, Info-NCE fails to boost the predictive performance by leveraging unlabeled data and the performance of Info-NCE becomes even worse than DNN. We conjecture that the unsupervised contrastive loss term introduces noise into the hidden representations due to the added Gaussian noise. We further analyze how different noise levels influence both the unsupervised contrastive loss and supervised contrastive loss in a case study presented in Section 4.4.


\begin{table*}
\centering
\caption{Results on XRMB and CelebA data sets. Notice that \method-s is identical to SupCon in multi-class setting (\eg, in XRMB data set).}
\begin{tabular}{|*{5}{c|}}
\hline - & \multicolumn{2}{c|}{XRMB} & \multicolumn{2}{c|}{CelebA}\\
\hline \textbf{Model}   & \textbf{F1 Score}     & \textbf{AUC}          & \textbf{F1 Score}     & \textbf{AUC} \\
\hline DNN              & 0.5600 $\pm$ 0.0106   & 0.9085 $\pm$ 0.0015   & 0.5474 $\pm$ 0.0130   & 0.7081 $\pm$ 0.0083 \\ 
\hline SupCon           & 0.5938 $\pm$ 0.0153   & 0.9207 $\pm$ 0.0062   & 0.5527 $\pm$ 0.0085   & 0.7166 $\pm$ 0.0088 \\
\hline CMC              & 0.6047 $\pm$ 0.0148   & 0.9336 $\pm$ 0.0019   & 0.5572 $\pm$ 0.0126   & 0.7213 $\pm$ 0.0132 \\
\hline MIB              & 0.5903 $\pm$ 0.0201   & 0.9204 $\pm$ 0.0073   & 0.5602 $\pm$ 0.0160   & 0.7222 $\pm$ 0.0075 \\
\hline C2AE             & 0.5850 $\pm$ 0.0161   & 0.9178 $\pm$ 0.0065   & 0.5726 $\pm$ 0.0125   & 0.7319 $\pm$ 0.0203 \\
\hline DeepMTMV         & 0.5898 $\pm$ 0.0155   & 0.9182 $\pm$ 0.0074   & 0.5621 $\pm$ 0.0098   & 0.7264 $\pm$ 0.0169 \\
\hline \method-s        & 0.5938 $\pm$ 0.0153   & 0.9207 $\pm$ 0.0062   & 0.5801 $\pm$ 0.0092   & 0.7497 $\pm$ 0.0074 \\
\hline \method-u        & 0.6169 $\pm$ 0.0128   & \textbf{0.9442 $\pm$ 0.0012}   & 0.5551 $\pm$ 0.0114   & 0.7307 $\pm$ 0.0044 \\
\hline \method          & \textbf{0.6180 $\pm$ 0.0147} & 0.9411 $\pm$ 0.0021 & \textbf{0.5907 $\pm$ 0.0121} & \textbf{0.7544 $\pm$ 0.0138} \\
\hline
\end{tabular}
\label{table_multi_view}
\vspace{-0.3cm}
\end{table*} 

\subsection{Multi-view Multi-label Setting}
\label{multi_view_multi_label}
In this subsection, we test the performance of our proposed method on two real-world data sets in a multi-view multi-label setting, including the XRMB data set and CelebA data set. In the experiments, we use Eq.~\ref{lu3} to compute the weighted unsupervised contrastive loss for ~\method~ and ~\method-u.

For the XRMB data set, we select the first 20 classes as the labels and randomly draw 2,500 samples from each class (50,000 samples in total). We sample 250 data points as our training set and the remaining 49,750 samples as our test set. The neural network architectures of DNN, CMC, SupCon, and our methods are the same, which is a three-layer fully-connected neural network. Two hyper-parameters $\alpha$ and $\beta$ for ~\method~ are $0.1$ and $0.001$, respectively; the hyper-parameter $\alpha$ for ~\method-${u}$ is $0.1$; and the hyper-parameter $\beta$ for ~\method-${s}$ is $0.01$. The batch size is 250, the number of epochs for our methods is 500 and the size of the negative set $|\mathcal{N}_i|$ is equal to 4,999. Table~\ref{table_multi_view} shows the performance of our proposed methods and state-of-the-art models. By observation, our proposed methods outperform all baselines in terms of F1 score and AUC. Similar to the observation on the Scene data set, DeepMTMV, MIB and C2AE only achieve better performance than DNN and behave worse than all contrastive learning based methods. Our conjecture is that these methods suffer a lot from insufficient label information.

For the CelebA data set, we randomly draw 500 samples as our training set and 49,500 samples as our test set. The number of binary labels is 40 and 9 binary labels out of 40 are labeled as positive on average. The neural network architectures of DNN, CMC, SupCon, and our methods are vgg-16~\cite{simonyan2014very}, consisting of thirteen convolutional layers, five max-pooling layers, and a three-layer fully-connected neural network. We set two hyper-parameters $\alpha=0.05$ and $\beta=0.1$ for ~\method, the hyper-parameter $\alpha=0.1$ for ~\method-${s}$, and the hyper-parameter $\beta=0.01$ for ~\method-${u}$. The batch size is 50, the number of epochs is 600 and the size of the negative set $|\mathcal{N}_i|$ is 249. Table~\ref{table_multi_view} shows the performance of our proposed methods and state-of-the-art baseline models. By observation, we could find that our proposed methods outperform all baselines in terms of both F1 score and AUC; DeepMTMV and MIB have slightly better predictive results than CMC and SupCon because they suffer from insufficient label information; C2AE outperforms all baselines except the our proposed method since it 
benefits from exploring the label dependency. As we mentioned in Section 3.2.2, due to the large number of unique label vectors in the CelebA data set, SupCon fails to maximize the similarity between the hidden representation of two samples with similar label vectors. Thus, Table~\ref{table_multi_view} shows that SupCon only performs slightly better than DNN but is worse than the rest of the algorithms.  Compared with SupCon, \method-s and \method~ improve the performance by 2.7\% and 3.8\% in terms of F1 score, respectively. 

\begin{figure}
\begin{center}
\begin{tabular}{cc}
\hspace{-0.5cm}
\includegraphics[width=0.50\linewidth]{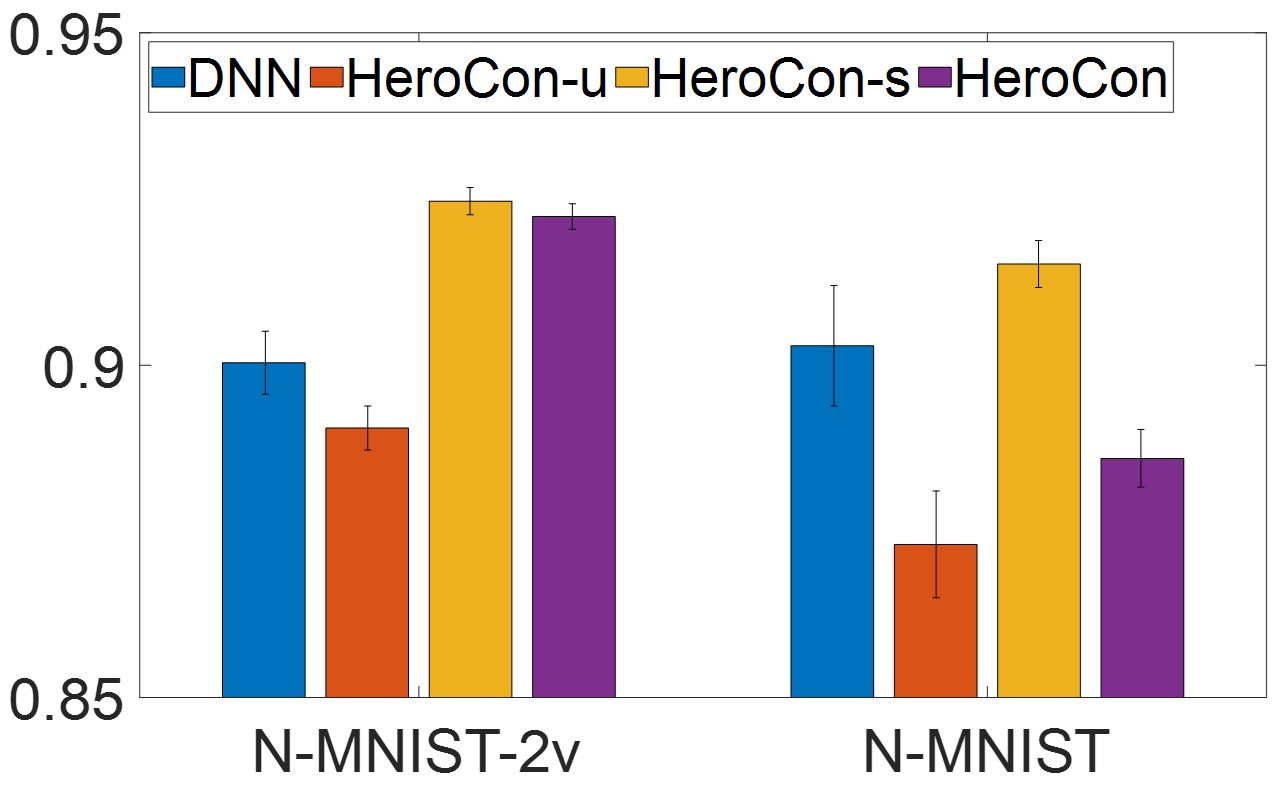}&
\includegraphics[width=0.50\linewidth]{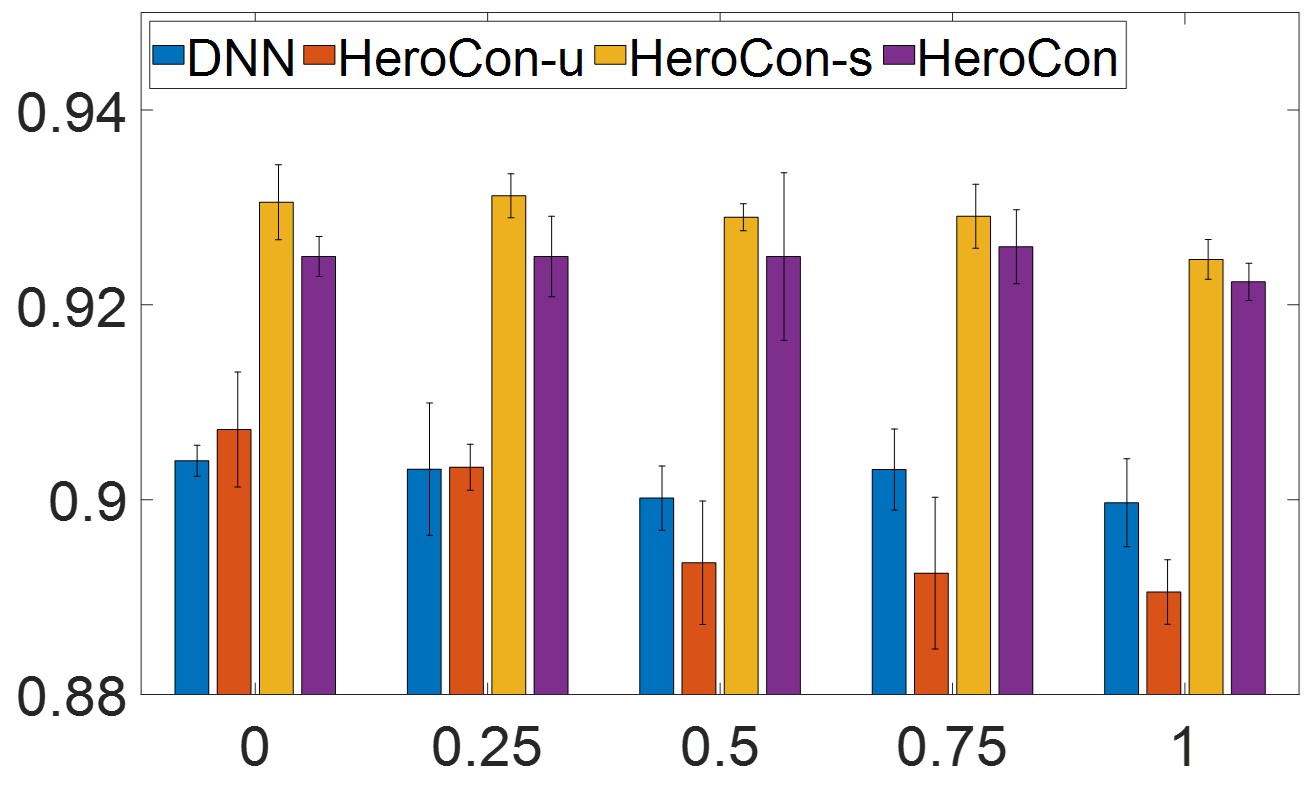} \\
\hspace{-0.5cm} (a) Results at noise level = 1 &
(b) Results at different noise \\
\hspace{-0.5cm} & 
levels on N-MNIST-2v\\
\end{tabular}
\end{center}
\caption{Performance comparison on N-MNIST data set with different levels of noise in terms of F1 score}
\label{fig_case_study}
\end{figure}

\subsection{Case Study}
In this subsection, we study how different noise level influences two weighted unsupervised contrastive loss term (\ie, Eq.~\ref{lu2} and Eq.~\ref{lu3}) and the weighted supervised contrastive loss term. We denote the single view Noisy MNIST as N-MNIST and two-view Noisy MNIST as N-MNIST-2v.
For a fair comparison, our data preprocessing consists of the following steps: (1) we generate the noise matrix $\mathcal{N}(\mu, \sigma)$ with the same shape as the entire data set, where $\mu$ and $\sigma$ are the mean and the standard deviation of the MNIST data set; (2) to generate different levels of noise, we sample different percentage of indices from the same noise matrix $\mathcal{N}(\mu, \sigma)$ (\eg, $0\%, 25\%, 50\%, 75\%$ and $100\%$)~\footnote{$75\%$ means that 75 percent of pixels are contaminated by random noise.} and add them to the original MNIST data to generate N-MNIST, which is also considered as the first view of Noisy-MNIST-2v; (3) to generate the second view for N-MNIST-2v, we repeat step 2 to generate the second noise matrix $\mathcal{N}_2(\mu, \sigma)$ in order to create the second view. 
In this case study, we set the number of epochs for our methods to be 500, the batch size to be 200 (the size of the entire labeled set), and the size of the negative set $|\mathcal{N}_i|$ to be 4,199. In addition, we use the same hyper-parameters for all of our methods, \eg, hyper-parameter $\alpha=0.5$ for \method-u~, $\beta=10$ for \method-s~ and $\alpha=0.5$ and $\beta=10$ for \method.

In Figure~\ref{fig_case_study} (a), the y-axis is the performance of 4 methods in terms of F1 score, and the left-hand side and the right-hand side of this figure show the performance of 4 methods for N-MNIST-2v and N-MNIST at noise level=1, respectively. In Figure~\ref{fig_case_study} (b), the x-axis is the level of the noise and the y-axis is the performance of 4 methods in terms of F1 score. By observation, we could find that in Figure~\ref{fig_case_study} (a), \method-s achieves the best performance and DNN remains the similar performance in both settings, while the F1 score of both \method-u and \method~ drop dramatically in the single view setting. \method-u for two views (\ie. Eq.~\ref{lu3}) performs better than \method-u for the single view (\ie. Eq.~\ref{lu2}). Our conjecture is that the goal of \method-u for the single view (\ie, Eq.~\ref{lu2}) is to maximize the mutual information between the hidden representation and the original input features, which introduces noise in the hidden representation if the input features contain a lot of noise (\eg, $100\%$ in current setting). 
By observation in Figure~\ref{fig_case_study} (b), in the two view setting, the performance of \method-s and \method~ is slightly influenced by the noise level and their performance does not change too much as the noise level increases. Therefore, we could make a conclusion that when the input data is contaminated by random noise,  \method-u for two views (\ie. Eq.~\ref{lu3}) has a better performance than \method-u for the single view (\ie. Eq.~\ref{lu2}), and \method-s and \method~ are robust enough to handle different levels of noise.

\begin{figure}
\begin{center}
\begin{tabular}{cc}
\includegraphics[width=0.49\linewidth]{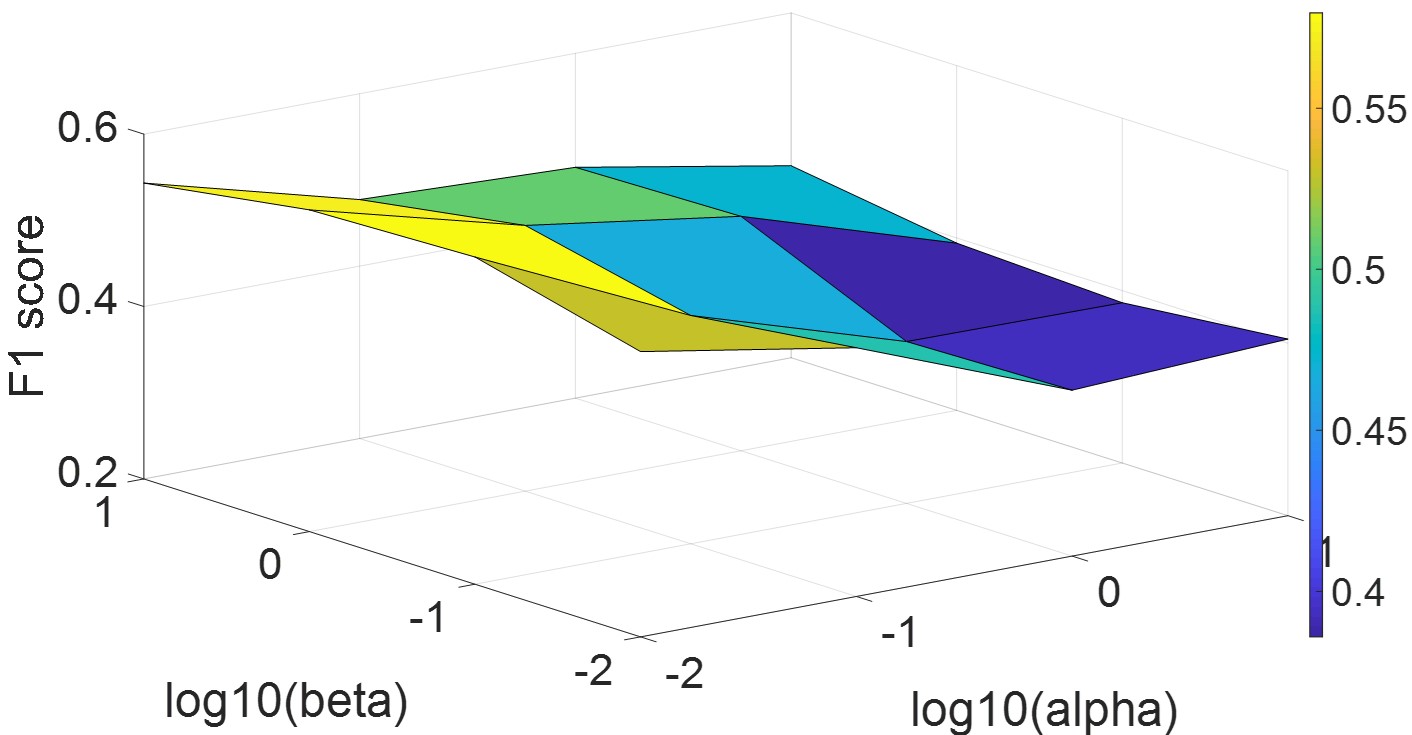} &
\includegraphics[width=0.49\linewidth]{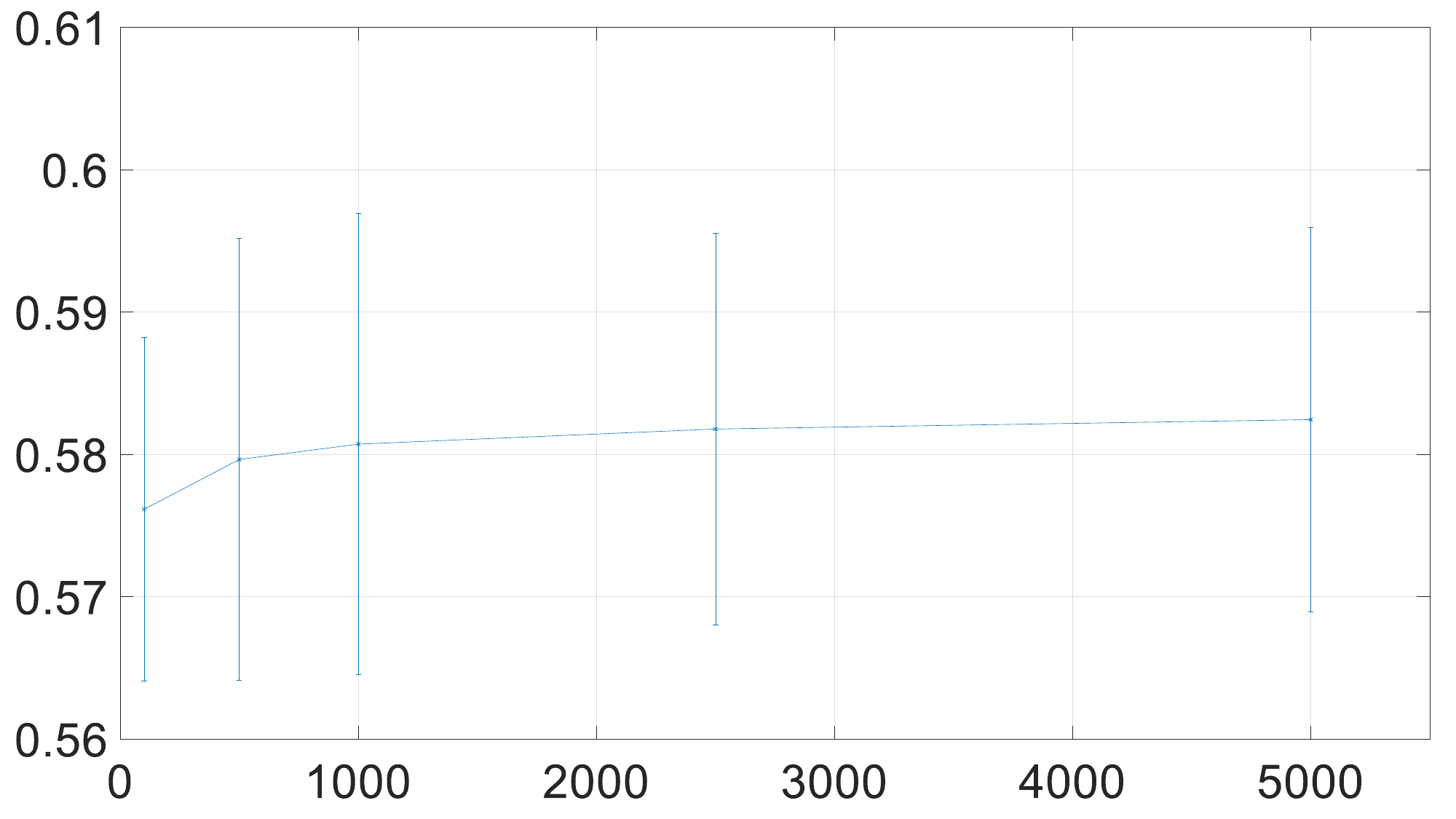} \\
(a) Parameter sensitivity &
(b) The size of negative set\\
(Best viewed in color)&
vs. F1 score\\
\end{tabular}
\end{center}
\caption{Parameter analysis on XRMB data set}
\label{fig_parameter_analysis}
\vspace{-0.3cm}
\end{figure}

\subsection{Parameter Analysis}
In this subsection, we analyze the parameter sensitivity of our proposed~\method~ algorithm on the XRMB data set, including $\alpha$, $\beta$ and the size of the negative set $|\mathcal{N}_i|$. In all experiments, we use 250 samples the training set and 49,750 samples as the test set; we set the batch size to be 250, the number of epochs to be 500, the learning rate to be 0.05; the optimizer is momentum stochastic gradient descent with Layer-wise Adaptive Rate Scaling scheduler (LARS)~\cite{you2017large}.
In the first experiment, we fix the size of the negative set $|\mathcal{N}_i|$ to be 4,999, adjust the value of both $\alpha$ and $\beta$ and record the F1 score of \method. The results are shown in Figure~\ref{fig_parameter_analysis} (a), where the x, y, z axes are the logarithm of $\alpha$ with base 10, the logarithm of $\beta$ with base 10 and the F1 score. By observation, a large value of $\beta$ and a small value of $\alpha$ usually leads to a better performance and it achieves the best performance at $\beta=10$ and $\alpha=0.01$ or $\log_{10}(\beta)=1$ and $\log_{10}(\alpha)=-2$.
As $\alpha$ and $\beta$ are the weight for the unsupervised contrastive learning loss and supervised contrastive learning loss, respectively, the large value of $\beta$ and small value of $\alpha$ with better performance suggests that \method~ mainly relies on the supervised contrastive regularization to improve the performance on the XRMB data set, because this term aims to bring the samples from the same class closer by leveraging the label information.

In the second experiment, we fix $\alpha=0.1$, $\beta=0.01$, and increase the size of the negative set $|\mathcal{N}_i|$ from 100 to 5,000. The experiments are repeated 5 times and the mean and standard deviation are reported. The results are shown in Figure~\ref{fig_parameter_analysis} (b), where the x-axis is the size of the negative set and the y-axis is the F1 score. By observation, we could see that the F1 score of \method~ increases as we increase the size of the negative set. Based on the theoretical analysis in Section 3.4 Lemma~\ref{lemma_u}, the mutual information between two samples is lower bounded by our proposed weighted unsupervised contrastive loss. As the size of the negative set becomes larger, the lower bound becomes tighter, which is demonstrated by Figure~\ref{fig_parameter_analysis} (b). However, as the size of the negative set increases, the computational cost also increases, which will be illustrated in the following subsection.

%% file: Conclusion.tex
\section{Conclusion}
In this paper, we propose \method\ - a deep contrastive learning framework for modeling complex heterogeneity. By proposing a weighed unsupervised contrastive loss to model the view heterogeneity, and a weighted supervised contrastive loss to model the label heterogeneity, our proposed framework is capable of handling multiple types of data heterogeneity in the presence of insufficient label information. We also provide theoretical analysis showing that the vanilla contrastive learning loss easily leads to the sub-optimal solution in the presence of false negative pairs, whereas the proposed weighted loss could automatically adjust the weight based on the similarity of the learned representations. In addition, we provide theoretical analysis to show that the proposed weighted supervised contrastive loss is the lower bound of the mutual information of two samples sharing similar label information and the weighted unsupervised contrastive loss is the lower bound of the mutual information between the hidden representations of two views of the same sample. 
The experimental results on real-world data sets demonstrate the effectiveness and efficiency of the proposed framework.

%% file: proof.tex
\section{Appendix}

\subsection{Efficiency Analysis}
\begin{figure}[H]
\begin{center}
\begin{tabular}{cc}
\includegraphics[width=0.49\linewidth]{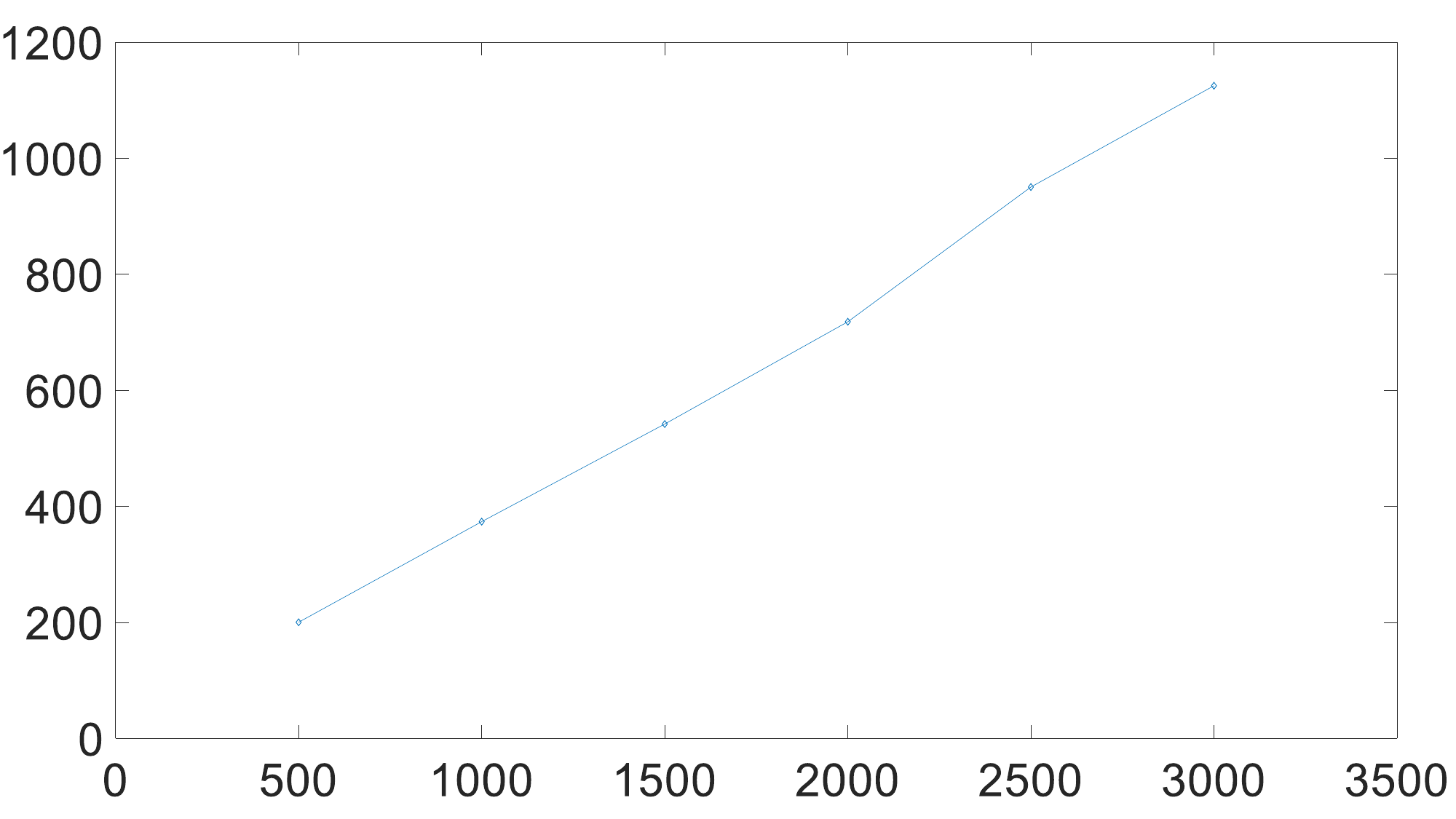} &
\includegraphics[width=0.49\linewidth]{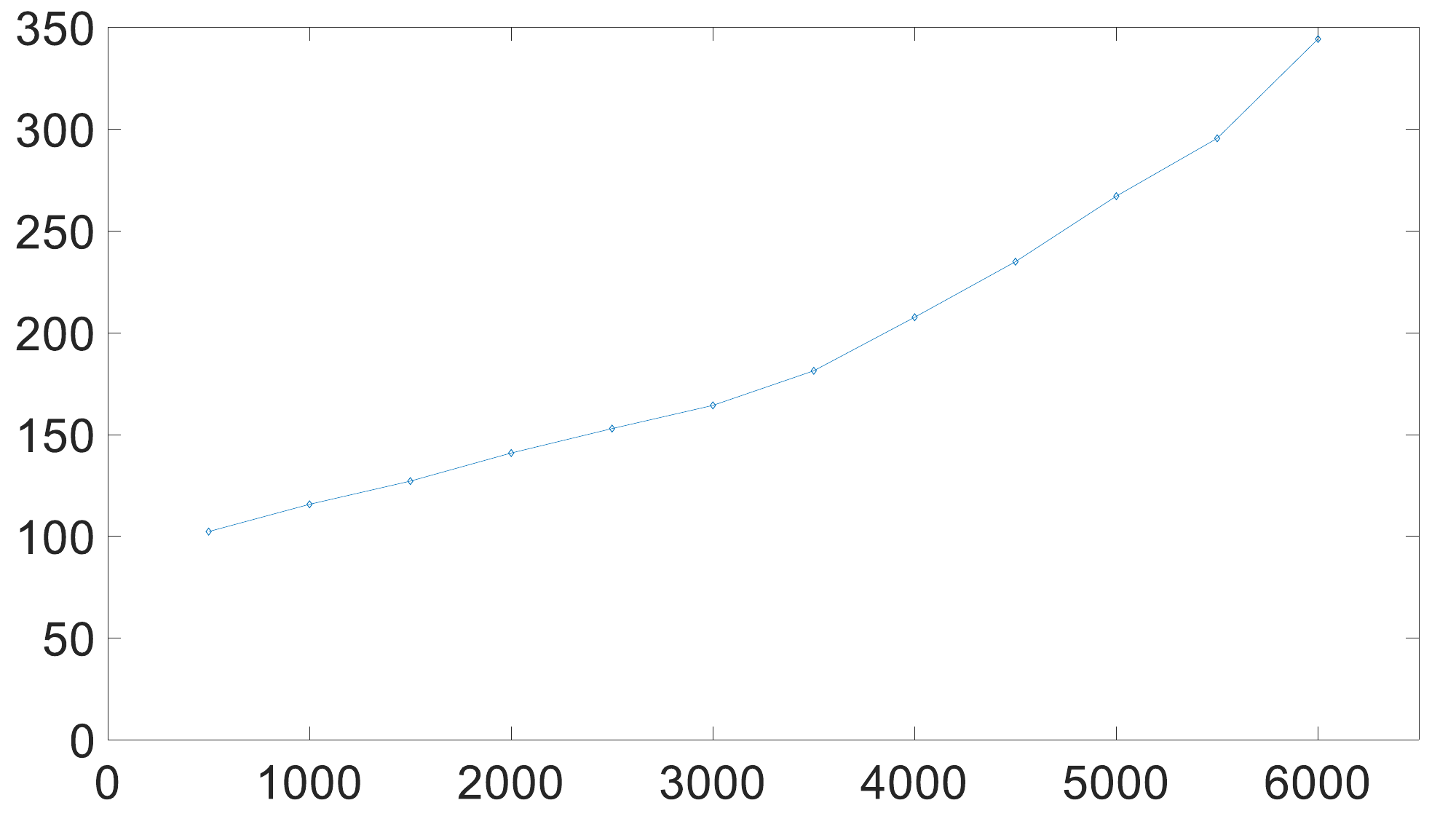} \\
(a) The size of training set &
(b) The size of negative set \\
vs. running time (in seconds) &
vs. running time (in seconds)\\
\end{tabular}
\end{center}
\caption{Efficiency analysis on XRMB data set}
\label{fig_efficiency_analysis}
\end{figure}

In this subsection, we analyze the efficiency of our proposed~\method~ algorithm with different sizes of the training set and different sizes of the negative set $|\mathcal{N}_i|$ on the XRMB data set. 
In the first experiment, we aim to see how the running time changes when we increase the size of the training set. First, we fix $\alpha=0.1$, $\beta=0.01$, the batch size to be 250, the number of epochs to be 500, and the size of negative set $|\mathcal{N}_i|$ to be 999. Then, we set the initial number of training samples to be 500, increase the size of the training set by 500 each time, and record the running time. The results are shown in Figure~\ref{fig_efficiency_analysis} (a), where the x-axis is the size of the training set or labeled set and the y-axis is the running time. By observation, we could see the running time is roughly linear to the size of the training set.

In the second experiment, we aim to see how the running time changes when we increase the size of the negative set. We first fix $\alpha=0.1$, $\beta=0.01$, the batch size to be 250, the number of epochs to be 500 and the number of training samples to be 250. Then, we set the initial size of the negative set $|\mathcal{N}_i|$ to be 500, increase the size of the negative set by 500 each time, and record the running time. The results are shown in Figure~\ref{fig_efficiency_analysis} (b), where the x-axis is the size of the negative set and the y-axis is the running time. By observation, we could see that the running time is roughly proportional to the square of the size of the negative set. Based on Eq.~\ref{lu3}, the negative set $|\mathcal{N}_i|$ is only involved in computing the weighted unsupervised contrastive loss. For each sample $\bm{X_i}$ drawn from the entire data set $\mathcal{D}$, we need to compute the similarity between this sample and the samples drawn from the negative set $|\mathcal{N}_i|$, which results in $O(n^2)$ time complexity, where $n$ is the size of entire data set $\mathcal{D}$. Combining the observations in Figure~\ref{fig_parameter_analysis} (b) and Figure~\ref{fig_efficiency_analysis} (b), we could see that there is a trade-off between the computational cost and the performance. The larger the size of the negative set, the higher performance the algorithm achieves but also the higher computational cost it suffers from.

\subsection{Theoretical Analysis}

\textbf{LEMMA 3.1.}
\textit{
Given the vanilla contrastive learning loss function $L_1$, if there exists one false negative sample in the batch during training, the contrastive learning loss will lead to a sub-optimal solution.}\\

\noindent\textbf{Proof:}
\textit{The unweighted contrastive learning loss could be written as follows:}
\begin{equation}
    \nonumber
    \begin{split}
        L_1 
        &= \frac{1}{n}\sum_{i=1}^{n}[\log (\frac{e^{\bm{Z_{i,1}}^T\bm{Z_{i,2}}/\tau}+\sum_{k\neq i} e^{\bm{Z_{i,1}}^T\bm{Z_{k,2}}/\tau}}{e^{\bm{Z_{i,1}}^T\bm{Z_{i, 2}}/\tau}})] \\
        &= \frac{1}{n}\sum_{i=1}^{n}[\log (e^{\bm{Z_{i,1}}^T\bm{Z_{i, 2}}/\tau} + \sum_{k\neq i} e^{\bm{Z_{i,1}}^T\bm{Z_{k,2}}/\tau})- \bm{Z_{i,1}}^T\bm{Z_{i, 2}/\tau}] \\
    \end{split}
\end{equation}
\textit{Taking the derivative of $L_1$ with respect to $\bm{Z}_{i,2}$, we have}

\begin{equation}
    \label{false_negative_11}
    \begin{split}
        \frac{\partial L_1}{\partial \bm{Z}_{i,2}} &= \frac{1}{n\tau} [\frac{e^{\bm{Z_{i,1}}^T\bm{Z_{i,2}}/\tau}\bm{Z_{i,1}} }{e^{\bm{Z_{i,1}}^T\bm{Z_{i,2}}/\tau} +\sum_{k\neq i} e^{\bm{Z_{i,1}}^T\bm{Z_{k,2}}/\tau}} - \bm{Z_{i, 1}}]\\
        &= \frac{1}{n\tau} [\frac{-\sum_{k\neq i} e^{\bm{Z_{i,1}}^T\bm{Z_{k,2}}/\tau}\bm{Z_{i, 1}}}{e^{\bm{Z_{i,1}}^T\bm{Z_{i,2}}/\tau} +\sum_{k\neq i} e^{\bm{Z_{i,1}}^T\bm{Z_{k,2}}/\tau}}]\\
    \end{split}
\end{equation}
\textit{By setting the gradient to be 0, we could get an equation in terms of the optimal representations $\bm{Z_{i,2}}^*$, $\bm{Z_{k,2}}^*$ and $\bm{Z_{i,1}}^*$.}
\begin{equation}
    \begin{split}
        \frac{\partial L_1}{\partial \bm{Z}_{i,2}}
        &= \frac{1}{n\tau} [\frac{-\sum_{k\neq i} e^{\bm{Z_{i,1}}^T\bm{Z_{k,2}}/\tau}\bm{Z_{i, 1}}}{e^{\bm{Z_{i,1}}^T\bm{Z_{i,2}}/\tau} +\sum_{k\neq i} e^{\bm{Z_{i,1}}^T\bm{Z_{k,2}}/\tau}}]=0\\
        &\sum_{k\neq i} e^{(\bm{Z_{i,1}}^*)^T\bm{Z_{k,2}}^*/\tau}\bm{Z_{i, 1}}^*=0
    \end{split}
    \label{false_negative_12}
\end{equation}
\textit{Since $\bm{Z_{i, 1}}^*$ is a non-zero vector, Eq.~\ref{false_negative_12} holds if we have $e^{(\bm{Z_{i,1}}^*)^T\bm{Z_{k,2}}^*/\tau}=0$ for all $k$. According to Definition~\ref{definition_2}, if ($\bm{Z_{i,1}}$, $\bm{Z_{k,2}}$) is a negative pair, then $e^{(\bm{Z_{i,1}}^*)^T\bm{Z_{k,2}}^*/\tau} \approx 0$ for some positive small values $\tau$ and thus Eq.~\ref{false_negative_12} holds.}
\textit{However, if there exists one false negative sample denoted as $\bm{Z_{j,2}}$ in the training batch, then $e^{(\bm{Z_{i,1}}^*)^T\bm{Z_{j,2}}^*/\tau} > 1$ for any false negative pair. This means that $\sum_{k\neq i} e^{(\bm{Z_{i,1}}^*)^T\bm{Z_{k,2}}^*/\tau}\bm{Z_{i, 2}}^* >1$, and thus we could not get the optimal solution for $\bm{Z_{i,2}}^*$, which completes the proof.}\\

\noindent\textbf{LEMMA 3.2.}
\textit{
Given two samples $\bm{X_i}$ and $\bm{X_j}$ from the same class drawn from the labeled set $\mathcal{L}$, we have $I(\bm{X_i}, \bm{X_j}) \geq -\frac{1}{\sigma} (L_s - N)$, where $I(\bm{X_i}, \bm{X_j})$ is the mutual information between $\bm{X_i}$ and $\bm{X_j}$,  $L_s$ is the supervised contrastive loss weighted by hamming distance measurement, 
$\sigma = 1- dist(\bm{Y_i}^\mathcal{L}, \bm{Y_j}^\mathcal{L})/c$,  which measures the ratio of two binary labels for two samples $\bm{X_i}$ and $\bm{X_j}$ having the same value,
and $N=\frac{1}{c}\sum_{a=1}^c \log(|\mathcal{N}^\mathcal{L}(a)|)$.}\\

\noindent\textbf{Proof:}
\textit{Following the theoretical analysis in~\cite{oord2018representation}, the optimal value of $f(\bm{S_i}, \bm{S_j})$ is given by $\frac{P(\bm{X_j}|\bm{X_i})}{P(\bm{X_j})}$.
Thus, the weighted supervised contrastive loss could be rewritten as follows:
\begin{equation}
    \notag
    \begin{split}
        L_s &= -\frac{1}{c}\sum_{a=1}^c \E_{\bm{X_i}, \bm{X_j} \in \mathcal{P}^\mathcal{L}(a)} [ \log \frac{\sigma f(\bm{S_i}, \bm{S_j})}{\sigma f(\bm{S_i}, \bm{S_j}) + \sum_{\bm{X_k} \in \mathcal{N}^\mathcal{L}(a)} \gamma f(\bm{S_i}, \bm{S_k})} ] \\
        &= \frac{1}{c}\sum_{a=1}^c \E_{\bm{X_i}, \bm{X_j} \in \mathcal{P}^\mathcal{L}(a)} [ \log \frac{\sigma f(\bm{S_i}, \bm{S_j}) + \sum_{\bm{X_k} \in \mathcal{N}^\mathcal{L}(a)} \gamma f(\bm{S_i}, \bm{S_k})}{\sigma f(\bm{S_i}, \bm{S_j})} ] \\
        &= \frac{1}{c}\sum_{a=1}^c \E_{\bm{X_i}, \bm{X_j} \in \mathcal{P}^\mathcal{L}(a)} [ \log \frac{\sigma \frac{P(\bm{X_j}|\bm{X_i})}{P(\bm{X_j})} + \sum_{\bm{X_k} \in \mathcal{N}^\mathcal{L}(a)} \gamma \frac{P(\bm{X_k}|\bm{X_i})}{P(\bm{X_k})}}{\sigma \frac{P(\bm{X_j}|\bm{X_i})}{P(\bm{X_j})}} ] \\
        &= \frac{1}{c}\sum_{a=1}^c \E_{\bm{X_i}, \bm{X_j} \in \mathcal{P}^\mathcal{L}(a)} [ \log( 1 + \frac{P(\bm{X_j})}{\sigma P(\bm{X_j}|\bm{X_i})} \sum_{\bm{X_k} \in \mathcal{N}^\mathcal{L}(a)} \gamma \frac{P(\bm{X_k}|\bm{X_i})}{P(\bm{X_k})})] \\
    \end{split}
\end{equation}
Since $(\bm{X_i^\mathcal{L}}, \bm{X_k}^\mathcal{L})$ is defined as a negative pair, it means that at least one binary label does not match for this negative pair. Therefore, we have $\gamma=dist(\bm{Y_i^\mathcal{L}}, \bm{Y_k}^\mathcal{L}) \in [1,c] $ for all negative pairs and $\sigma \in [\frac{1}{c}, 1]$ for all positive pairs with hamming distance measurement, which leads to  $\frac{P(\bm{X_j})}{\sigma P(\bm{X_j}|\bm{X_i})} \geq \frac{P(\bm{X_j})}{P(\bm{X_j}|\bm{X_i})}$ and $\gamma \frac{P(\bm{X_k}|\bm{X_i})}{P(\bm{X_k})} \geq \frac{P(\bm{X_k}|\bm{X_i})}{P(\bm{X_k})}$.
Thus, we have
\begin{equation}
    \notag
    \begin{split}
        L_s &\geq \frac{1}{c}\sum_{a=1}^c \E_{\bm{X_i}, \bm{X_j} \in \mathcal{P}^\mathcal{L}(a)} [ \log( 1 + \frac{P(\bm{X_j})}{ P(\bm{X_j}|\bm{X_i})} \sum_{\bm{X_k} \in \mathcal{N}^\mathcal{L}(a)} \frac{P(\bm{X_k}|\bm{X_i})}{P(\bm{X_k})})] \\
        &\approx \frac{1}{c}\sum_{a=1}^c \E_{\bm{X_i}, \bm{X_j} \in \mathcal{P}^\mathcal{L}(a)} [ \log( 1 + \frac{P(\bm{X_j})}{ P(\bm{X_j}|\bm{X_i})}  (|\mathcal{N}^\mathcal{L}(a)| \E_{X_k} \frac{P(\bm{X_k}|\bm{X_i})}{P(\bm{X_k})})] \\
        &= \frac{1}{c}\sum_{a=1}^c \E_{\bm{X_i}, \bm{X_j} \in \mathcal{P}^\mathcal{L}(a)} [ \log( 1 + \frac{P(\bm{X_j})}{ P(\bm{X_j}|\bm{X_i})}  |\mathcal{N}^\mathcal{L}(a)|)] \\
        &\geq \frac{1}{c}\sum_{a=1}^c \E_{\bm{X_i}, \bm{X_j} \in \mathcal{P}^\mathcal{L}(a)} [ \log(\frac{P(\bm{X_j})}{ P(\bm{X_j}|\bm{X_i})}) + \log(|\mathcal{N}^\mathcal{L}(a)|)] \\
        &= -(1- dist(\bm{Y_i}^\mathcal{L}, \bm{Y_j}^\mathcal{L})/c)I(\bm{X_i}, \bm{X_j}) + \frac{1}{c}\sum_{a=1}^c \log(|\mathcal{N}^\mathcal{L}(a)|)] \\
        &= -\sigma I(\bm{X_i}, \bm{X_j}) + N \\ 
    \end{split}
\end{equation}
where $I(\bm{X_i}, \bm{X_j})=\E_{\bm{X_i}, \bm{X_j}} [\log(\frac{P(\bm{X_j})}{ P(\bm{X_j}|\bm{X_i})})]$,  $|\mathcal{N}^\mathcal{L}(a)|$ is the number of negative pairs for the  $a^{th}$ label, $N=\frac{1}{c}\sum_{a=1}^c \log(|\mathcal{N}^\mathcal{L}(a)|)$ and  $\sigma = 1- dist(\bm{Y_i}^\mathcal{L}, \bm{Y_j}^\mathcal{L})/c$. Finally, we have $I(\bm{X_i}, \bm{X_j}) \geq -\frac{1}{\sigma} (L_s - N)$, which completes the proof.}\\

\noindent\textbf{LEMMA 3.3.}
\textit{
Given a sample $\bm{X_i}$ drawn from the entire set $\mathcal{D}$, we have $I(\bm{X_{i,1}}, \bm{X_{i,2}}) \geq - L_u + \log(|\mathcal{N}^\mathcal{D}_i|)$, where $I(\bm{X_{i,1}}, \bm{X_{i,2}})$ is the mutual information between $\bm{X_{i,1}}$ and $\bm{X_{i,2}}$, $L_u$ is the unsupervised contrastive loss weighted by $g(\bm{Z_{i,1}}, \bm{Z_{k,j}})$ and $|\mathcal{N}^\mathcal{D}_i|$ is the size of negative set.}\\

\noindent\textbf{Proof:}
\textit{Similar to the theoretical analysis in Lemma~\ref{lemma_s}, the optimal value of $f(\bm{X_{i,1}}, \bm{X_{i,2}})$ is given by $\frac{P(\bm{X_{i,2}}|\bm{X_{i,1}})}{P(\bm{X_{i,2}})}$.
Thus, the weighted unsupervised contrastive loss could be rewritten as follows:
\begin{equation}
    \notag
    \begin{split}
         L_u & = -\E_{X_i\in \mathcal{D}}[  \log \frac{f(\bm{Z_{i,1}}, \bm{Z_{i,2}})}{f(\bm{Z_{i,1}}, \bm{Z_{i,2}}) + \sum_{\bm{X_{k}} \in \mathcal{N}_{i}^\mathcal{D}} g(\bm{Z_{i,1}}, \bm{Z_{k,j}}) f(\bm{Z_{i,1}}, \bm{Z_{k,j}})} ] \\
         & = \E_{X_i\in \mathcal{D}} [ \log( 1 + \frac{P(\bm{X_{i,2}})}{P(\bm{X_{i,2}}|\bm{X_{i,1}})} \sum_{\bm{X_{k}} \in \mathcal{N}^\mathcal{D}_i}  \frac{g(\bm{Z_{i,1}}, \bm{Z_{k,j}})P(\bm{X_{k,j}}|\bm{X_{i,1}})}{P(\bm{X_{k,j}})})] \\
    \end{split}
\end{equation}
Notice that $g(\bm{Z_{i,1}}, \bm{Z_{k,j}}) \in [1, e^2]$,
so $g(\bm{Z_{i,1}}, \bm{Z_{k,j}}) \frac{P(\bm{X_{k,j}}|\bm{X_{i,1}})}{P(\bm{X_{k,j}})} \geq \frac{P(\bm{X_{k,j}}|\bm{X_{i,1}})}{P(\bm{X_{k,j}})}$.
Similarly, we have 
\begin{equation}
    \notag
    \begin{split}
         L_u & \geq \E_{X_i\in \mathcal{D}} [ \log( 1 + \frac{P(\bm{X_{i,2}})}{P(\bm{X_{i,2}}|\bm{X_{i,1}})} \sum_{\bm{X_{k}} \in \mathcal{N}^\mathcal{D}_i} \frac{P(\bm{X_{k,j}}|\bm{X_{i,1}})}{P(\bm{X_{k,j}})})] \\
         &\approx \E_{X_i\in \mathcal{D}} [ \log( 1 + \frac{P(\bm{X_{i,2}})}{P(\bm{X_{i,2}}|\bm{X_{i,1}})}  |\mathcal{N}^\mathcal{D}_i| \E_{\bm{X_{k}} \in \mathcal{N}^\mathcal{D}_i} \frac{P(\bm{X_{k,j}}|\bm{X_{i,1}})}{P(\bm{X_{k,j}})})] \\
         &=  \E_{X_i\in \mathcal{D}} [ \log( 1 + \frac{P(\bm{X_{i,2}})}{P(\bm{X_{i,2}}|\bm{X_{i,1}})}  |\mathcal{N}^\mathcal{D}_i|)] \\
         &\geq  \E_{X_i\in \mathcal{D}} [ \log(\frac{P(\bm{X_{i,2}})}{P(\bm{X_{i,2}}|\bm{X_{i,1}})}) +\log(|\mathcal{N}^\mathcal{D}_i|)] \\
         &\geq  -I(\bm{X_{i,1}}, \bm{X_{i,2}}) + \log(|\mathcal{N}^\mathcal{D}_i|) \\
    \end{split}
\end{equation}
Finally, we have $I(\bm{X_{i,1}}, \bm{X_{i,2}}) \geq - L_u + \log(|\mathcal{N}^\mathcal{D}_i|)$, which completes the proof.}

%% file: main.bbl

\begin{thebibliography}{51}


\ifx \showCODEN    \undefined \def \showCODEN     #1{\unskip}     \fi
\ifx \showDOI      \undefined \def \showDOI       #1{#1}\fi
\ifx \showISBNx    \undefined \def \showISBNx     #1{\unskip}     \fi
\ifx \showISBNxiii \undefined \def \showISBNxiii  #1{\unskip}     \fi
\ifx \showISSN     \undefined \def \showISSN      #1{\unskip}     \fi
\ifx \showLCCN     \undefined \def \showLCCN      #1{\unskip}     \fi
\ifx \shownote     \undefined \def \shownote      #1{#1}          \fi
\ifx \showarticletitle \undefined \def \showarticletitle #1{#1}   \fi
\ifx \showURL      \undefined \def \showURL       {\relax}        \fi
\providecommand\bibfield[2]{#2}
\providecommand\bibinfo[2]{#2}
\providecommand\natexlab[1]{#1}
\providecommand\showeprint[2][]{arXiv:#2}

\bibitem[Akaho(2006)]%
        {DBLP:journals/corr/abs-cs-0609071}
\bibfield{author}{\bibinfo{person}{Shotaro Akaho}.}
  \bibinfo{year}{2006}\natexlab{}.
\newblock \showarticletitle{A kernel method for canonical correlation
  analysis}.
\newblock \bibinfo{journal}{\emph{CoRR}}  \bibinfo{volume}{abs/cs/0609071}
  (\bibinfo{year}{2006}).
\newblock


\bibitem[Basu et~al\mbox{.}(2017)]%
        {basu2017learning}
\bibfield{author}{\bibinfo{person}{Saikat Basu}, \bibinfo{person}{Manohar
  Karki}, \bibinfo{person}{Sangram Ganguly}, \bibinfo{person}{Robert DiBiano},
  \bibinfo{person}{Supratik Mukhopadhyay}, \bibinfo{person}{Shreekant Gayaka},
  \bibinfo{person}{Rajgopal Kannan}, {and} \bibinfo{person}{Ramakrishna~R.
  Nemani}.} \bibinfo{year}{2017}\natexlab{}.
\newblock \showarticletitle{Learning Sparse Feature Representations Using
  Probabilistic Quadtrees and Deep Belief Nets}.
\newblock \bibinfo{journal}{\emph{Neural Process. Lett.}} \bibinfo{volume}{45},
  \bibinfo{number}{3} (\bibinfo{year}{2017}), \bibinfo{pages}{855--867}.
\newblock


\bibitem[Blum and Mitchell(1998)]%
        {DBLP:conf/colt/BlumM98}
\bibfield{author}{\bibinfo{person}{Avrim Blum} {and} \bibinfo{person}{Tom~M.
  Mitchell}.} \bibinfo{year}{1998}\natexlab{}.
\newblock \showarticletitle{Combining Labeled and Unlabeled Data with
  Co-Training}. In \bibinfo{booktitle}{\emph{Proceedings of the Eleventh Annual
  Conference on Computational Learning Theory, {COLT} 1998}}.
  \bibinfo{publisher}{{ACM}}, \bibinfo{pages}{92--100}.
\newblock


\bibitem[Boutell et~al\mbox{.}(2004)]%
        {boutell2004learning}
\bibfield{author}{\bibinfo{person}{Matthew~R. Boutell}, \bibinfo{person}{Jiebo
  Luo}, \bibinfo{person}{Xipeng Shen}, {and} \bibinfo{person}{Christopher~M.
  Brown}.} \bibinfo{year}{2004}\natexlab{}.
\newblock \showarticletitle{Learning multi-label scene classification}.
\newblock \bibinfo{journal}{\emph{Pattern Recognit.}} \bibinfo{volume}{37},
  \bibinfo{number}{9} (\bibinfo{year}{2004}), \bibinfo{pages}{1757--1771}.
\newblock


\bibitem[Chen et~al\mbox{.}(2020)]%
        {chen2020simple}
\bibfield{author}{\bibinfo{person}{Ting Chen}, \bibinfo{person}{Simon
  Kornblith}, \bibinfo{person}{Mohammad Norouzi}, {and}
  \bibinfo{person}{Geoffrey~E. Hinton}.} \bibinfo{year}{2020}\natexlab{}.
\newblock \showarticletitle{A Simple Framework for Contrastive Learning of
  Visual Representations}. In \bibinfo{booktitle}{\emph{Proceedings of the 37th
  International Conference on Machine Learning, {ICML} 2020}},
  Vol.~\bibinfo{volume}{119}. \bibinfo{publisher}{{PMLR}},
  \bibinfo{pages}{1597--1607}.
\newblock


\bibitem[Chuang et~al\mbox{.}(2020)]%
        {chuang2020debiased}
\bibfield{author}{\bibinfo{person}{Ching{-}Yao Chuang}, \bibinfo{person}{Joshua
  Robinson}, \bibinfo{person}{Yen{-}Chen Lin}, \bibinfo{person}{Antonio
  Torralba}, {and} \bibinfo{person}{Stefanie Jegelka}.}
  \bibinfo{year}{2020}\natexlab{}.
\newblock \showarticletitle{Debiased Contrastive Learning}. In
  \bibinfo{booktitle}{\emph{Advances in Annual Conference on Neural Information
  Processing Systems 2020}}.
\newblock


\bibitem[Federici et~al\mbox{.}(2020)]%
        {federici2020learning}
\bibfield{author}{\bibinfo{person}{Marco Federici}, \bibinfo{person}{Anjan
  Dutta}, \bibinfo{person}{Patrick Forr{\'{e}}}, \bibinfo{person}{Nate
  Kushman}, {and} \bibinfo{person}{Zeynep Akata}.}
  \bibinfo{year}{2020}\natexlab{}.
\newblock \showarticletitle{Learning Robust Representations via Multi-View
  Information Bottleneck}.
\newblock  (\bibinfo{year}{2020}).
\newblock


\bibitem[Feng et~al\mbox{.}(2022)]%
        {feng2022adversarial}
\bibfield{author}{\bibinfo{person}{Shengyu Feng}, \bibinfo{person}{Baoyu Jing},
  \bibinfo{person}{Yada Zhu}, {and} \bibinfo{person}{Hanghang Tong}.}
  \bibinfo{year}{2022}\natexlab{}.
\newblock \showarticletitle{Adversarial Graph Contrastive Learning with
  Information Regularization}. In \bibinfo{booktitle}{\emph{{WWW} '22: The
  {ACM} Web Conference 2022}}. \bibinfo{publisher}{{ACM}},
  \bibinfo{pages}{1362--1371}.
\newblock


\bibitem[Fu et~al\mbox{.}(2020)]%
        {DBLP:conf/cikm/FuXLTH20}
\bibfield{author}{\bibinfo{person}{Dongqi Fu}, \bibinfo{person}{Zhe Xu},
  \bibinfo{person}{Bo Li}, \bibinfo{person}{Hanghang Tong}, {and}
  \bibinfo{person}{Jingrui He}.} \bibinfo{year}{2020}\natexlab{}.
\newblock \showarticletitle{A View-Adversarial Framework for Multi-View Network
  Embedding}. In \bibinfo{booktitle}{\emph{{CIKM} '20: The 29th {ACM}
  International Conference on Information and Knowledge Management, Virtual
  Event, Ireland, October 19-23, 2020}}. \bibinfo{publisher}{{ACM}},
  \bibinfo{pages}{2025--2028}.
\newblock


\bibitem[He and Lawrence(2011)]%
        {he2011graphbased}
\bibfield{author}{\bibinfo{person}{Jingrui He} {and} \bibinfo{person}{Rick
  Lawrence}.} \bibinfo{year}{2011}\natexlab{}.
\newblock \showarticletitle{A Graphbased Framework for Multi-Task Multi-View
  Learning}. In \bibinfo{booktitle}{\emph{Proceedings of the 28th International
  Conference on Machine Learning, {ICML} 2011}}.
  \bibinfo{publisher}{Omnipress}, \bibinfo{pages}{25--32}.
\newblock


\bibitem[Hong et~al\mbox{.}(2013)]%
        {DBLP:conf/iccv/HongMPT13}
\bibfield{author}{\bibinfo{person}{Zhibin Hong}, \bibinfo{person}{Xue Mei},
  \bibinfo{person}{Danil~V. Prokhorov}, {and} \bibinfo{person}{Dacheng Tao}.}
  \bibinfo{year}{2013}\natexlab{}.
\newblock \showarticletitle{Tracking via Robust Multi-task Multi-view Joint
  Sparse Representation}. In \bibinfo{booktitle}{\emph{{IEEE} International
  Conference on Computer Vision, {ICCV} 2013}}. \bibinfo{publisher}{{IEEE}
  Computer Society}, \bibinfo{pages}{649--656}.
\newblock


\bibitem[Huang et~al\mbox{.}(2014)]%
        {DBLP:conf/aaai/HuangGZ14}
\bibfield{author}{\bibinfo{person}{Sheng{-}Jun Huang}, \bibinfo{person}{Wei
  Gao}, {and} \bibinfo{person}{Zhi{-}Hua Zhou}.}
  \bibinfo{year}{2014}\natexlab{}.
\newblock \showarticletitle{Fast Multi-Instance Multi-Label Learning}. In
  \bibinfo{booktitle}{\emph{Proceedings of the Twenty-Eighth {AAAI} Conference
  on Artificial Intelligence, July 27 -31, 2014, Qu{\'{e}}bec City,
  Qu{\'{e}}bec, Canada}}. \bibinfo{publisher}{{AAAI}},
  \bibinfo{pages}{1868--1874}.
\newblock


\bibitem[Huo et~al\mbox{.}(2020)]%
        {huo2020heterogeneous}
\bibfield{author}{\bibinfo{person}{Xinyue Huo}, \bibinfo{person}{Lingxi Xie},
  \bibinfo{person}{Longhui Wei}, \bibinfo{person}{Xiaopeng Zhang},
  \bibinfo{person}{Hao Li}, \bibinfo{person}{Zijie Yang},
  \bibinfo{person}{Wengang Zhou}, \bibinfo{person}{Houqiang Li}, {and}
  \bibinfo{person}{Qi Tian}.} \bibinfo{year}{2020}\natexlab{}.
\newblock \showarticletitle{Heterogeneous contrastive learning: Encoding
  spatial information for compact visual representations}.
\newblock \bibinfo{journal}{\emph{arXiv preprint arXiv:2011.09941}}
  (\bibinfo{year}{2020}).
\newblock


\bibitem[Jing et~al\mbox{.}(2021a)]%
        {jing2021hdmi}
\bibfield{author}{\bibinfo{person}{Baoyu Jing}, \bibinfo{person}{Chanyoung
  Park}, {and} \bibinfo{person}{Hanghang Tong}.}
  \bibinfo{year}{2021}\natexlab{a}.
\newblock \showarticletitle{{HDMI:} High-order Deep Multiplex Infomax}. In
  \bibinfo{booktitle}{\emph{{WWW} '21: The Web Conference 2021}}.
  \bibinfo{publisher}{{ACM} / {IW3C2}}, \bibinfo{pages}{2414--2424}.
\newblock


\bibitem[Jing et~al\mbox{.}(2021b)]%
        {jing2021graph}
\bibfield{author}{\bibinfo{person}{Baoyu Jing}, \bibinfo{person}{Yuejia Xiang},
  \bibinfo{person}{Xi Chen}, \bibinfo{person}{Yu Chen}, {and}
  \bibinfo{person}{Hanghang Tong}.} \bibinfo{year}{2021}\natexlab{b}.
\newblock \showarticletitle{Graph-MVP: Multi-View Prototypical Contrastive
  Learning for Multiplex Graphs}.
\newblock \bibinfo{journal}{\emph{arXiv preprint arXiv:2109.03560}}
  (\bibinfo{year}{2021}).
\newblock


\bibitem[Khosla et~al\mbox{.}(2020)]%
        {khosla2020supervised}
\bibfield{author}{\bibinfo{person}{Prannay Khosla}, \bibinfo{person}{Piotr
  Teterwak}, \bibinfo{person}{Chen Wang}, \bibinfo{person}{Aaron Sarna},
  \bibinfo{person}{Yonglong Tian}, \bibinfo{person}{Phillip Isola},
  \bibinfo{person}{Aaron Maschinot}, \bibinfo{person}{Ce Liu}, {and}
  \bibinfo{person}{Dilip Krishnan}.} \bibinfo{year}{2020}\natexlab{}.
\newblock \showarticletitle{Supervised Contrastive Learning}. In
  \bibinfo{booktitle}{\emph{Advances in Annual Conference on Neural Information
  Processing Systems 2020}}.
\newblock


\bibitem[Kim and Xing(2010)]%
        {DBLP:conf/icml/KimX10}
\bibfield{author}{\bibinfo{person}{Seyoung Kim} {and} \bibinfo{person}{Eric~P.
  Xing}.} \bibinfo{year}{2010}\natexlab{}.
\newblock \showarticletitle{Tree-Guided Group Lasso for Multi-Task Regression
  with Structured Sparsity}. In \bibinfo{booktitle}{\emph{Proceedings of the
  27th International Conference on Machine Learning, 2010}}.
  \bibinfo{publisher}{Omnipress}, \bibinfo{pages}{543--550}.
\newblock


\bibitem[Lanckriet et~al\mbox{.}(2002)]%
        {DBLP:conf/icml/LanckrietCBGJ02}
\bibfield{author}{\bibinfo{person}{Gert R.~G. Lanckriet},
  \bibinfo{person}{Nello Cristianini}, \bibinfo{person}{Peter~L. Bartlett},
  \bibinfo{person}{Laurent~El Ghaoui}, {and} \bibinfo{person}{Michael~I.
  Jordan}.} \bibinfo{year}{2002}\natexlab{}.
\newblock \showarticletitle{Learning the Kernel Matrix with Semi-Definite
  Programming}. In \bibinfo{booktitle}{\emph{Proceedings of the Nineteenth
  International Conference {(ICML} 2002)}}. \bibinfo{publisher}{Morgan
  Kaufmann}, \bibinfo{pages}{323--330}.
\newblock


\bibitem[LeCun et~al\mbox{.}(1998)]%
        {lecun1998gradient}
\bibfield{author}{\bibinfo{person}{Yann LeCun}, \bibinfo{person}{L{\'e}on
  Bottou}, \bibinfo{person}{Yoshua Bengio}, {and} \bibinfo{person}{Patrick
  Haffner}.} \bibinfo{year}{1998}\natexlab{}.
\newblock \showarticletitle{Gradient-based learning applied to document
  recognition}.
\newblock \bibinfo{journal}{\emph{Proc. IEEE}} \bibinfo{volume}{86},
  \bibinfo{number}{11} (\bibinfo{year}{1998}), \bibinfo{pages}{2278--2324}.
\newblock


\bibitem[Li et~al\mbox{.}(2022)]%
        {li2022graph}
\bibfield{author}{\bibinfo{person}{Bolian Li}, \bibinfo{person}{Baoyu Jing},
  {and} \bibinfo{person}{Hanghang Tong}.} \bibinfo{year}{2022}\natexlab{}.
\newblock \showarticletitle{Graph Communal Contrastive Learning}. In
  \bibinfo{booktitle}{\emph{{WWW} '22: The {ACM} Web Conference 2022}}.
  \bibinfo{publisher}{{ACM}}, \bibinfo{pages}{1203--1213}.
\newblock


\bibitem[Liu et~al\mbox{.}(2015)]%
        {liu2015deep}
\bibfield{author}{\bibinfo{person}{Ziwei Liu}, \bibinfo{person}{Ping Luo},
  \bibinfo{person}{Xiaogang Wang}, {and} \bibinfo{person}{Xiaoou Tang}.}
  \bibinfo{year}{2015}\natexlab{}.
\newblock \showarticletitle{Deep Learning Face Attributes in the Wild}. In
  \bibinfo{booktitle}{\emph{2015 {IEEE} International Conference on Computer
  Vision, {ICCV} 2015}}. \bibinfo{publisher}{{IEEE} Computer Society},
  \bibinfo{pages}{3730--3738}.
\newblock


\bibitem[Lu et~al\mbox{.}(2017)]%
        {DBLP:conf/cvpr/LuKZCJF17}
\bibfield{author}{\bibinfo{person}{Yongxi Lu}, \bibinfo{person}{Abhishek
  Kumar}, \bibinfo{person}{Shuangfei Zhai}, \bibinfo{person}{Yu Cheng},
  \bibinfo{person}{Tara Javidi}, {and} \bibinfo{person}{Rogerio Feris}.}
  \bibinfo{year}{2017}\natexlab{}.
\newblock \showarticletitle{Fully-adaptive feature sharing in multi-task
  networks with applications in person attribute classification}. In
  \bibinfo{booktitle}{\emph{Proceedings of the IEEE conference on computer
  vision and pattern recognition}}. \bibinfo{pages}{5334--5343}.
\newblock


\bibitem[Luo et~al\mbox{.}(2013)]%
        {DBLP:conf/aaai/LuoTXLX13}
\bibfield{author}{\bibinfo{person}{Yong Luo}, \bibinfo{person}{Dacheng Tao},
  \bibinfo{person}{Chang Xu}, \bibinfo{person}{Dongchen Li}, {and}
  \bibinfo{person}{Chao Xu}.} \bibinfo{year}{2013}\natexlab{}.
\newblock \showarticletitle{Vector-Valued Multi-View Semi-Supervsed Learning
  for Multi-Label Image Classification}. In
  \bibinfo{booktitle}{\emph{Proceedings of the Twenty-Seventh Conference on
  Artificial Intelligence}}. \bibinfo{publisher}{{AAAI}}.
\newblock


\bibitem[Mao et~al\mbox{.}(2014)]%
        {DBLP:journals/corr/MaoXYWY14}
\bibfield{author}{\bibinfo{person}{Junhua Mao}, \bibinfo{person}{Wei Xu},
  \bibinfo{person}{Yi Yang}, \bibinfo{person}{Jiang Wang}, {and}
  \bibinfo{person}{Alan~L. Yuille}.} \bibinfo{year}{2014}\natexlab{}.
\newblock \showarticletitle{Explain Images with Multimodal Recurrent Neural
  Networks}.
\newblock \bibinfo{journal}{\emph{CoRR}}  \bibinfo{volume}{abs/1410.1090}
  (\bibinfo{year}{2014}).
\newblock


\bibitem[Misra et~al\mbox{.}(2016)]%
        {DBLP:conf/cvpr/MisraSGH16}
\bibfield{author}{\bibinfo{person}{Ishan Misra}, \bibinfo{person}{Abhinav
  Shrivastava}, \bibinfo{person}{Abhinav Gupta}, {and} \bibinfo{person}{Martial
  Hebert}.} \bibinfo{year}{2016}\natexlab{}.
\newblock \showarticletitle{Cross-Stitch Networks for Multi-task Learning}. In
  \bibinfo{booktitle}{\emph{2016 {IEEE} Conference on Computer Vision and
  Pattern Recognition}}. \bibinfo{publisher}{{IEEE} Computer Society},
  \bibinfo{pages}{3994--4003}.
\newblock


\bibitem[Nie et~al\mbox{.}(2018)]%
        {DBLP:journals/tip/NieCLL18}
\bibfield{author}{\bibinfo{person}{Feiping Nie}, \bibinfo{person}{Guohao Cai},
  \bibinfo{person}{Jing Li}, {and} \bibinfo{person}{Xuelong Li}.}
  \bibinfo{year}{2018}\natexlab{}.
\newblock \showarticletitle{Auto-Weighted Multi-View Learning for Image
  Clustering and Semi-Supervised Classification}.
\newblock \bibinfo{journal}{\emph{{IEEE} Trans. Image Process.}}
  \bibinfo{volume}{27}, \bibinfo{number}{3} (\bibinfo{year}{2018}),
  \bibinfo{pages}{1501--1511}.
\newblock


\bibitem[Nigam and Ghani(2000)]%
        {nigam2000analyzing}
\bibfield{author}{\bibinfo{person}{Kamal Nigam} {and} \bibinfo{person}{Rayid
  Ghani}.} \bibinfo{year}{2000}\natexlab{}.
\newblock \showarticletitle{Analyzing the Effectiveness and Applicability of
  Co-training}. In \bibinfo{booktitle}{\emph{Proceedings of the 2000 {ACM}
  {CIKM} International Conference on Information and Knowledge Management,
  2000}}. \bibinfo{publisher}{{ACM}}, \bibinfo{pages}{86--93}.
\newblock


\bibitem[Pupo et~al\mbox{.}(2015)]%
        {DBLP:journals/ijon/PupoMV15}
\bibfield{author}{\bibinfo{person}{Oscar Gabriel~Reyes Pupo},
  \bibinfo{person}{Carlos Morell}, {and} \bibinfo{person}{Sebasti{\'{a}}n
  Ventura}.} \bibinfo{year}{2015}\natexlab{}.
\newblock \showarticletitle{Scalable extensions of the ReliefF algorithm for
  weighting and selecting features on the multi-label learning context}.
\newblock \bibinfo{journal}{\emph{Neurocomputing}}  \bibinfo{volume}{161}
  (\bibinfo{year}{2015}), \bibinfo{pages}{168--182}.
\newblock


\bibitem[Simonyan and Zisserman(2015)]%
        {simonyan2014very}
\bibfield{author}{\bibinfo{person}{Karen Simonyan} {and}
  \bibinfo{person}{Andrew Zisserman}.} \bibinfo{year}{2015}\natexlab{}.
\newblock \showarticletitle{Very Deep Convolutional Networks for Large-Scale
  Image Recognition}. In \bibinfo{booktitle}{\emph{3rd International Conference
  on Learning Representations, {ICLR} 2015}}.
\newblock


\bibitem[Sohn(2016)]%
        {DBLP:conf/nips/Sohn16}
\bibfield{author}{\bibinfo{person}{Kihyuk Sohn}.}
  \bibinfo{year}{2016}\natexlab{}.
\newblock \showarticletitle{Improved Deep Metric Learning with Multi-class
  N-pair Loss Objective}. In \bibinfo{booktitle}{\emph{Advances in Annual
  Conference on Neural Information Processing Systems 2016}}.
  \bibinfo{pages}{1849--1857}.
\newblock


\bibitem[Song and Ermon(2020)]%
        {song2020multi}
\bibfield{author}{\bibinfo{person}{Jiaming Song} {and} \bibinfo{person}{Stefano
  Ermon}.} \bibinfo{year}{2020}\natexlab{}.
\newblock \showarticletitle{Multi-label Contrastive Predictive Coding}. In
  \bibinfo{booktitle}{\emph{Advances in Annual Conference on Neural Information
  Processing Systems 2020}}.
\newblock


\bibitem[Tian et~al\mbox{.}(2020)]%
        {tian2019contrastive}
\bibfield{author}{\bibinfo{person}{Yonglong Tian}, \bibinfo{person}{Dilip
  Krishnan}, {and} \bibinfo{person}{Phillip Isola}.}
  \bibinfo{year}{2020}\natexlab{}.
\newblock \showarticletitle{Contrastive Multiview Coding}. In
  \bibinfo{booktitle}{\emph{Computer Vision - {ECCV} 2020 - 16th European
  Conference}}, Vol.~\bibinfo{volume}{12356}. \bibinfo{publisher}{Springer},
  \bibinfo{pages}{776--794}.
\newblock


\bibitem[van~den Oord et~al\mbox{.}(2018)]%
        {oord2018representation}
\bibfield{author}{\bibinfo{person}{A{\"{a}}ron van~den Oord},
  \bibinfo{person}{Yazhe Li}, {and} \bibinfo{person}{Oriol Vinyals}.}
  \bibinfo{year}{2018}\natexlab{}.
\newblock \showarticletitle{Representation Learning with Contrastive Predictive
  Coding}.
\newblock \bibinfo{journal}{\emph{CoRR}}  \bibinfo{volume}{abs/1807.03748}
  (\bibinfo{year}{2018}).
\newblock


\bibitem[Wang et~al\mbox{.}(2015)]%
        {WangALB15}
\bibfield{author}{\bibinfo{person}{Weiran Wang}, \bibinfo{person}{Raman Arora},
  \bibinfo{person}{Karen Livescu}, {and} \bibinfo{person}{Jeff~A. Bilmes}.}
  \bibinfo{year}{2015}\natexlab{}.
\newblock \showarticletitle{On Deep Multi-View Representation Learning}. In
  \bibinfo{booktitle}{\emph{Proceedings of the 32nd International Conference on
  Machine Learning, {ICML} 2015}}, Vol.~\bibinfo{volume}{37}.
  \bibinfo{publisher}{JMLR.org}, \bibinfo{pages}{1083--1092}.
\newblock


\bibitem[Westbury(1994)]%
        {westbury1994x}
\bibfield{author}{\bibinfo{person}{JR Westbury}.}
  \bibinfo{year}{1994}\natexlab{}.
\newblock \showarticletitle{X-ray microbeam speech production database user’s
  handbook: Madison}.
\newblock \bibinfo{journal}{\emph{WI: Waisman Center, University of Wisconsin}}
  (\bibinfo{year}{1994}).
\newblock


\bibitem[Wu and He(2021)]%
        {wu2021indirect}
\bibfield{author}{\bibinfo{person}{Jun Wu} {and} \bibinfo{person}{Jingrui He}.}
  \bibinfo{year}{2021}\natexlab{}.
\newblock \showarticletitle{Indirect Invisible Poisoning Attacks on Domain
  Adaptation}. In \bibinfo{booktitle}{\emph{Proceedings of the 27th ACM
  SIGKDD}}. \bibinfo{pages}{1852--1862}.
\newblock


\bibitem[Xu et~al\mbox{.}(2015a)]%
        {DBLP:journals/pami/XuTX15}
\bibfield{author}{\bibinfo{person}{Chang Xu}, \bibinfo{person}{Dacheng Tao},
  {and} \bibinfo{person}{Chao Xu}.} \bibinfo{year}{2015}\natexlab{a}.
\newblock \showarticletitle{Multi-View Intact Space Learning}.
\newblock \bibinfo{journal}{\emph{{IEEE} Trans. Pattern Anal. Mach. Intell.}}
  \bibinfo{volume}{37}, \bibinfo{number}{12} (\bibinfo{year}{2015}),
  \bibinfo{pages}{2531--2544}.
\newblock


\bibitem[Xu et~al\mbox{.}(2015b)]%
        {DBLP:journals/tip/XuT015}
\bibfield{author}{\bibinfo{person}{Chang Xu}, \bibinfo{person}{Dacheng Tao},
  {and} \bibinfo{person}{Chao Xu}.} \bibinfo{year}{2015}\natexlab{b}.
\newblock \showarticletitle{Multi-View Learning With Incomplete Views}.
\newblock \bibinfo{journal}{\emph{{IEEE} Trans. Image Process.}}
  \bibinfo{volume}{24}, \bibinfo{number}{12} (\bibinfo{year}{2015}),
  \bibinfo{pages}{5812--5825}.
\newblock


\bibitem[Xu et~al\mbox{.}(2016)]%
        {DBLP:conf/kdd/XuT016}
\bibfield{author}{\bibinfo{person}{Chang Xu}, \bibinfo{person}{Dacheng Tao},
  {and} \bibinfo{person}{Chao Xu}.} \bibinfo{year}{2016}\natexlab{}.
\newblock \showarticletitle{Robust Extreme Multi-label Learning}. In
  \bibinfo{booktitle}{\emph{Proceedings of the 22nd {ACM} {SIGKDD} 2016}}.
  \bibinfo{publisher}{{ACM}}, \bibinfo{pages}{1275--1284}.
\newblock


\bibitem[Yang et~al\mbox{.}(2015)]%
        {DBLP:journals/ijcv/YangMNCH15}
\bibfield{author}{\bibinfo{person}{Yi Yang}, \bibinfo{person}{Zhigang Ma},
  \bibinfo{person}{Feiping Nie}, \bibinfo{person}{Xiaojun Chang}, {and}
  \bibinfo{person}{Alexander~G. Hauptmann}.} \bibinfo{year}{2015}\natexlab{}.
\newblock \showarticletitle{Multi-Class Active Learning by Uncertainty Sampling
  with Diversity Maximization}.
\newblock \bibinfo{journal}{\emph{Int. J. Comput. Vis.}} \bibinfo{volume}{113},
  \bibinfo{number}{2} (\bibinfo{year}{2015}), \bibinfo{pages}{113--127}.
\newblock


\bibitem[Yeh et~al\mbox{.}(2017)]%
        {yeh2017learning}
\bibfield{author}{\bibinfo{person}{Chih-Kuan Yeh}, \bibinfo{person}{Wei-Chieh
  Wu}, \bibinfo{person}{Wei-Jen Ko}, {and} \bibinfo{person}{Yu-Chiang~Frank
  Wang}.} \bibinfo{year}{2017}\natexlab{}.
\newblock \showarticletitle{Learning deep latent space for multi-label
  classification}. In \bibinfo{booktitle}{\emph{Thirty-first AAAI conference on
  artificial intelligence}}.
\newblock


\bibitem[You et~al\mbox{.}(2017)]%
        {you2017large}
\bibfield{author}{\bibinfo{person}{Yang You}, \bibinfo{person}{Igor Gitman},
  {and} \bibinfo{person}{Boris Ginsburg}.} \bibinfo{year}{2017}\natexlab{}.
\newblock \showarticletitle{Large batch training of convolutional networks}.
\newblock \bibinfo{journal}{\emph{arXiv preprint arXiv:1708.03888}}
  (\bibinfo{year}{2017}).
\newblock


\bibitem[Zhang and Zhang(2010)]%
        {zhang2010multi}
\bibfield{author}{\bibinfo{person}{Min{-}Ling Zhang} {and} \bibinfo{person}{Kun
  Zhang}.} \bibinfo{year}{2010}\natexlab{}.
\newblock \showarticletitle{Multi-label learning by exploiting label
  dependency}. In \bibinfo{booktitle}{\emph{Proceedings of the 16th {ACM}
  {SIGKDD} 2010}}. \bibinfo{publisher}{{ACM}}, \bibinfo{pages}{999--1008}.
\newblock


\bibitem[Zheng et~al\mbox{.}(2019)]%
        {DBLP:conf/sdm/ZhengCH19}
\bibfield{author}{\bibinfo{person}{Lecheng Zheng}, \bibinfo{person}{Yu Cheng},
  {and} \bibinfo{person}{Jingrui He}.} \bibinfo{year}{2019}\natexlab{}.
\newblock \showarticletitle{Deep Multimodality Model for Multi-task Multi-view
  Learning}. In \bibinfo{booktitle}{\emph{Proceedings of the 2019 {SIAM}
  International Conference on Data Mining, {SDM} 2019}}.
  \bibinfo{publisher}{{SIAM}}, \bibinfo{pages}{10--18}.
\newblock


\bibitem[Zheng et~al\mbox{.}(2021a)]%
        {zheng2021deep}
\bibfield{author}{\bibinfo{person}{Lecheng Zheng}, \bibinfo{person}{Yu Cheng},
  \bibinfo{person}{Hongxia Yang}, \bibinfo{person}{Nan Cao}, {and}
  \bibinfo{person}{Jingrui He}.} \bibinfo{year}{2021}\natexlab{a}.
\newblock \showarticletitle{Deep Co-Attention Network for Multi-View Subspace
  Learning}. In \bibinfo{booktitle}{\emph{Proceedings of the Web Conference
  2021}}. \bibinfo{pages}{1528--1539}.
\newblock


\bibitem[Zheng et~al\mbox{.}(2021b)]%
        {zheng2021tackling}
\bibfield{author}{\bibinfo{person}{Lecheng Zheng}, \bibinfo{person}{Dongqi Fu},
  {and} \bibinfo{person}{Jingrui He}.} \bibinfo{year}{2021}\natexlab{b}.
\newblock \showarticletitle{Tackling Oversmoothing of GNNs with Contrastive
  Learning}.
\newblock \bibinfo{journal}{\emph{arXiv preprint arXiv:2110.13798}}
  (\bibinfo{year}{2021}).
\newblock


\bibitem[Zhou and Burges(2007)]%
        {DBLP:conf/icml/ZhouB07}
\bibfield{author}{\bibinfo{person}{Dengyong Zhou} {and}
  \bibinfo{person}{Christopher J.~C. Burges}.} \bibinfo{year}{2007}\natexlab{}.
\newblock \showarticletitle{Spectral clustering and transductive learning with
  multiple views}. In \bibinfo{booktitle}{\emph{Proceedings of the
  Twenty-Fourth International Conference {(ICML} 2007)}},
  Vol.~\bibinfo{volume}{227}. \bibinfo{publisher}{{ACM}},
  \bibinfo{pages}{1159--1166}.
\newblock


\bibitem[Zhou et~al\mbox{.}(2015)]%
        {zhou2015muvir}
\bibfield{author}{\bibinfo{person}{Dawei Zhou}, \bibinfo{person}{Jingrui He},
  \bibinfo{person}{K.~Sel{\c{c}}uk Candan}, {and} \bibinfo{person}{Hasan
  Davulcu}.} \bibinfo{year}{2015}\natexlab{}.
\newblock \showarticletitle{{MUVIR:} Multi-View Rare Category Detection}. In
  \bibinfo{booktitle}{\emph{Proceedings of the Twenty-Fourth International
  Joint Conference on Artificial Intelligence, {IJCAI} 2015}}.
  \bibinfo{publisher}{{AAAI} Press}, \bibinfo{pages}{4098--4104}.
\newblock


\bibitem[Zhou et~al\mbox{.}(2020)]%
        {zhou2020domain}
\bibfield{author}{\bibinfo{person}{Dawei Zhou}, \bibinfo{person}{Lecheng
  Zheng}, \bibinfo{person}{Yada Zhu}, \bibinfo{person}{Jianbo Li}, {and}
  \bibinfo{person}{Jingrui He}.} \bibinfo{year}{2020}\natexlab{}.
\newblock \showarticletitle{Domain Adaptive Multi-Modality Neural Attention
  Network for Financial Forecasting}. In \bibinfo{booktitle}{\emph{{WWW} '20:
  The Web Conference 2020}}. \bibinfo{publisher}{{ACM} / {IW3C2}},
  \bibinfo{pages}{2230--2240}.
\newblock


\bibitem[Zhou et~al\mbox{.}(2011)]%
        {DBLP:conf/nips/ZhouCY11}
\bibfield{author}{\bibinfo{person}{Jiayu Zhou}, \bibinfo{person}{Jianhui Chen},
  {and} \bibinfo{person}{Jieping Ye}.} \bibinfo{year}{2011}\natexlab{}.
\newblock \showarticletitle{Clustered Multi-Task Learning Via Alternating
  Structure Optimization}. In \bibinfo{booktitle}{\emph{Advances in 25th Annual
  Conference on Neural Information Processing Systems 2011.}}
  \bibinfo{pages}{702--710}.
\newblock


\bibitem[Zhu et~al\mbox{.}(2018)]%
        {DBLP:journals/tkde/ZhuKZ18}
\bibfield{author}{\bibinfo{person}{Yue Zhu}, \bibinfo{person}{James~T. Kwok},
  {and} \bibinfo{person}{Zhi{-}Hua Zhou}.} \bibinfo{year}{2018}\natexlab{}.
\newblock \showarticletitle{Multi-Label Learning with Global and Local Label
  Correlation}.
\newblock \bibinfo{journal}{\emph{{IEEE} Trans. Knowl. Data Eng.}}
  \bibinfo{volume}{30}, \bibinfo{number}{6} (\bibinfo{year}{2018}),
  \bibinfo{pages}{1081--1094}.
\newblock


\end{thebibliography}
